\pdfoutput=1
\documentclass[11pt]{article}

\usepackage[final]{coling}

\usepackage{times}
\usepackage{latexsym}

\usepackage[T1]{fontenc}
\usepackage[utf8]{inputenc}

\usepackage{microtype}

\usepackage{inconsolata}

\usepackage{graphicx}
\usepackage{subfig}
\usepackage{booktabs}
\usepackage{multirow}
\usepackage{tabularx}
\usepackage{pifont}

\newcommand{\ours}{\textsc{AIGT}}
\newcommand{\datas}{\textsc{STabs}}
\definecolor{darkgreen}{rgb}{0.0, 0.5, 0.0}
\usepackage{tikz}
\usepackage{pgfplots}
\usepackage{xcolor}
\definecolor{mycolor1}{HTML}{BEB8DC}
\definecolor{mycolor2}{HTML}{FFBE7A}
\definecolor{mycolor3}{HTML}{82B0D2}
\usepackage{listings}
\title{~\ours{}: AI Generative Table Based on Prompt}

\author{
	Mingming Zhang$^{1*}$ , Zhiqing Xiao$^{1*}$, Guoshan Lu$^{1*}$, Sai Wu$^{1}$, \\ 
    \textbf{Weiqiang Wang$^{2}$, Xing Fu$^{2}$,
    Can Yi$^{2}$, Junbo Zhao$^{1}$\footnotemark[2]}\\
	$^1$  Zhejiang University, Hangzhou, China\\
	$^2$ Ant Group, Hangzhou, China\\
    \texttt{\{mmz,zhiqing.xiao,luguoshan,wusai,j.zhao\}@zju.edu.cn}\\
    \texttt{\{weiqiang.wwq,zicai.fx,yican.yc\}@antgroup.com}
}

\begin{document}
\maketitle
% \footnotetext[1]{means both authors contributed equally to this research.}
\renewcommand{\thefootnote}{\fnsymbol{footnote}}
\footnotetext[1]{Indicates equal contribution.}
\footnotetext[2]{Corresponding author.}
\renewcommand{\thefootnote}{\arabic{footnote}}

\begin{abstract}

Tabular data, which accounts for over 80\% of enterprise data assets, is vital in various fields. With growing concerns about privacy protection and data-sharing restrictions, generating high-quality synthetic tabular data has become essential. 
Recent advancements show that large language models (LLMs) can effectively generate realistic tabular data by leveraging semantic information and overcoming the challenges of high-dimensional data that arise from one-hot encoding. However, current methods do not fully utilize the rich information available in tables.
To address this, we introduce \textbf{\underline{AI}} \textbf{\underline{G}}enerative \textbf{\underline{T}}able (\ours{}) based on prompt enhancement, a novel approach that utilizes metadata information, such as table descriptions and schemas, as prompts to generate ultra-high-quality synthetic data. To overcome the token limit constraints of LLMs, we propose long-token partitioning algorithms that enable \ours{} to model tables of any scale. 
~\ours{} achieves state-of-the-art performance on 14 out of 20 public datasets and two real industry datasets within the Alipay risk control system. 
%The source code, data, and other artifacts are available in  supplementary materials.

\end{abstract}

\section{Introduction}
% Synthetic data is crucial for safeguarding privacy, enhancing machine learning generalization, and improving performance in small-sample scenarios. Given the prevalence of tabular data in practical applications, its synthesis is essential yet challenging due to specificity, impurities, class imbalances, and privacy concerns.
% Given the prevalence of tabular data in practical applications, synthesizing tabular data is an extremely crucial task that not only ensures privacy protection but also significantly boosts the generalization capabilities of machine learning models and enhances performance in situations with limited samples. However, this task faces unique challenges such as specificity, impurities, class imbalances, and privacy concerns.
Given the prevalence of tabular data in practical applications, synthesizing high-quality tabular data is an essential task. It ensures privacy protection, enhances the generalization capabilities of machine learning models, and boosts performance in scenarios with limited samples. However, this task is fraught with unique challenges, including specificity, impurities, class imbalances, and privacy concerns.

% Most tabular data synthesis work~\cite{pr1,tablegan,pr3-ctgan,pr4-survey} uses generative models or statistical methods, which often lose textual information and struggle with complex feature relationships. Recent attempts using language models~\cite{borisov2023great,zhang2023taptap} incorporate feature names for better contextual learning. GReaT~\cite{borisov2023great} and TapTap~\cite{zhang2023taptap} have shown improvements but they are still constrained by token limits and incomplete utilization of tabular information.
Traditional approaches to tabular data synthesis, such as generative models and statistical methods, often lose textual information and struggle with capturing complex feature relationships~\cite{pr1,tablegan,pr3-ctgan,pr4-survey}. Recent efforts have explored the use of language models to incorporate feature names for better contextual learning. 
For instance, GReaT~\cite{borisov2023great} made the pioneering attempt to generate tabular data using large language models (LLMs), achieving satisfactory synthetic data. TapTap~\cite{zhang2023taptap} further enhanced the quality of synthetic data through table pre-training and applied it to data augmentation in tabular data for improved performance in prediction tasks, achieving state-of-the-art (SOTA) results. However, these methods are constrained by token limits, limiting their ability to handle arbitrarily wide tables. Additionally, they fail to fully utilize crucial tabular elements such as headers and column names, resulting in incomplete information integration.

Recently, Artificial Intelligence
Generation and Creation (AIGC) technologies, including large-scale language models and automatic image generation techniques, are driving progress in AI's capabilities for content creation and data generation. Prompt Learning, as an innovative approach, aligns the pre-training of language models with specific downstream tasks. By using well-designed prompts, which include task descriptions, input data, background information and output indicators, create a template that directs the model's attention and thus improves the accuracy and relevance of the generated context.

% Prompt learning is essential in text generation, as it aligns the pre-training of language models with specific downstream tasks. By using well-designed prompts, tasks are reframed to align with language modeling objectives, allowing models to generate relevant and coherent text with minimal task-specific data. These prompts, which include task descriptions, input data, background information, and output indicators, create a template that directs the model's attention and thus improves the accuracy and relevance of the generated text.

% Similarly, 
AIGC is more broadly applied to the generation of multimedia content, such as images, videos, and audio. Combined with Prompt Learning, AIGC systems can generate content more accurately according to user instructions or needs. However, The application of AIGC technology in the generation of tabular data is currently relatively limited. Although work on table generation based on large-scale language models has begun to explore, these efforts have mostly utilized only the cell values in the tables, without fully leveraging the more comprehensive information contained in the tabular data.
We recognize that table metadata, such as headers and column names, provides valuable context often overlooked in analysis. Inspired by AIGC and Prompt learning, we propose a prompt-enhanced model that leverages this metadata to improve performance.
Therefore,
we propose \textbf{\underline{AI}} \textbf{\underline{G}}enerative \textbf{\underline{T}}able (\ours{}) technique based on prompt enhancement, 
which leverages metadata  to enhance the quality and relevance of table generation. 

Methods based on language models inherently face limitations in generating long sequences of tokens, which is why previous approaches like GReaT and TapTap struggled to model tables with a large number of columns—something very common in real-world industry data. To address this, we developed a long-token partitioning algorithm specifically adapted for AIGT. This approach allows AIGT to effectively handle data synthesis tasks for tables of any size, overcoming the token length constraints. We further demonstrate AIGT's effectiveness in real-world scenarios such as Alipay's risk control system for commercial credit and merchant fraud detection.
% Methods based on language models inherently have limitations when it comes to generating long tokens, which is why previous approaches like GReat and TapTap were unable to model tables with a large number of column names. However, such tables are very common in real-world industry data. To address this issue, we designed a long-token partitioning algorithm perfectly adapted to AIGT. This algorithm allows AIGT to support data synthesis tasks for tables of any size. We also demonstrate the effectiveness of AIGT in Alipay risk control system for commercial credit and merchant fraud scenarios.
%这个是原来的描述
% To address the token limitation challenge in language models, we designed a long-token partitioning algorithm perfectly adapted to AIGT. This algorithm allows AIGT to support data synthesis tasks for tables of any size. We also demonstrate the effectiveness of AIGT in Alipay risk control system for commercial credit and merchant fraud scenarios. 
% We demonstrate its effectiveness in Alipay's risk control system for commercial credit and merchant fraud scenarios.
% It uses prompt-enhanced training strategies and a long-token partitioning algorithm to overcome token limitations, allowing support for any table size. 
% We demonstrate its effectiveness in Alipay's risk control system for commercial credit and merchant fraud scenarios.

% We demonstrate the effectiveness of AIGT in Alipay risk control system for commercial credit and merchant fraud scenarios. 
Our key contributions are as follows: (1) \textbf{A Prompt-Enhanced LM for Tabular Data Synthesis:} We utilize the metadata of tables to construct prompt-enhanced language model and design a series of training techniques to enhance the capability of table generation. (2) \textbf{Scalability:} Our proposed partitioning algorithm enables LM-based generation methods to be scaled to tables of any size, overcoming the token limit constraints of LM methods. (3) \textbf{Superior Performance:} We achieve state-of-the-art results on 14 out of 20 academic datasets and two real-world industry datasets from Alipay.
% , and we provide a user-friendly Python implementation, accessible via pip installation.
% \begin{enumerate}
%     \item We propose a prompt-enhanced approach for tabular data synthesis, which includes designing a training strategy to improve tabular data training on language models.
%     \item We create a long-token partitioning algorithm that supports data synthesis for tables of any size.
%     \item 
%     \item We provide a user-friendly Python toolbox of AIGT, which is accessible via pip installation and achieves state-of-the-art performance on 20 academic datasets and 2 practical industrial scenarios.
% \end{enumerate}
% Our contributions are: i) Proposing a prompt-enhanced approach for tabular data synthesis, achieving state-of-the-art performance on 20 academic datasets and two industrial scenarios. ii) Designing a prompt-enhanced training strategy for better tabular training on language models. iii) Creating a long-token partitioning algorithm to support any table size. iv) Providing a user-friendly Python implementation of AIGT, accessible via pip installation.

\section{Related Works}
In this section, we provide a brief background on prompt engineering and tabular data synthesis approaches.

\begin{figure*}[!ht]
\centering
\includegraphics[width=\linewidth]{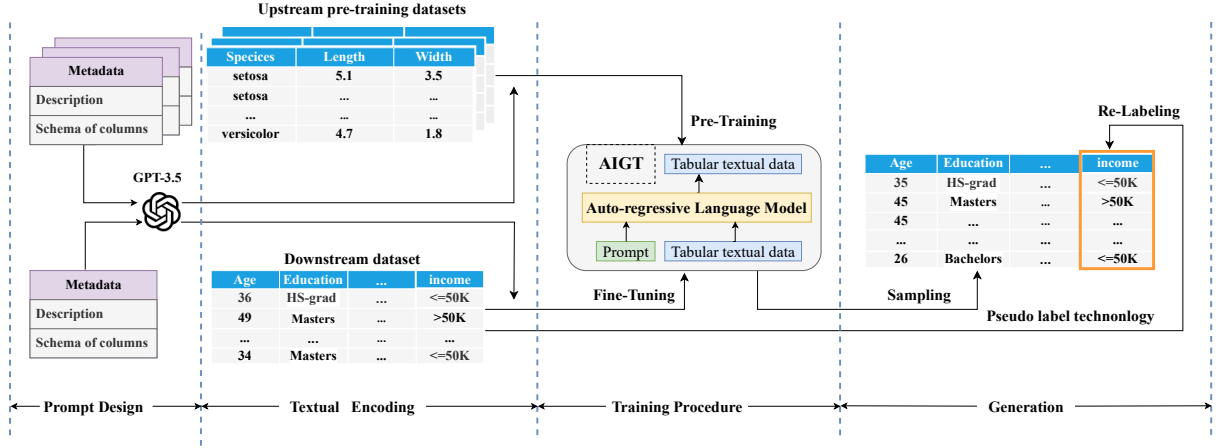}
\caption{The architecture of the proposed \ours{}. Firstly, AIGT utilizes the collected pre-trained corpus for pre-training; Then perform fine-tuning training on each downstream table to learn the complex relationships between features; Finally, based on the trained language model model, sample rows can be composed.}
\label{Fig: overview}
\end{figure*}

\subsection{Prompt Engineering}
% Prompt engineering has recently gained significant traction as a technique for improving the performance and usability of AI models, particularly in the domain of natural language processing (NLP). The concept of a "prompt" as described by \cite{prompt}, involves providing natural language instructions or commands to guide the AI model in completing tasks. This method has shown considerable benefits, including enhanced model effectiveness, reduced training time and cost, and improved interpretability and controllability. More notable works include \cite{jiang-etal-2020-know,liu23acm,brown2020nips}.
Prompt engineering has recently garnered significant attention as a technique for enhancing the performance and usability of AI models, especially in natural language processing (NLP). As described by \cite{prompt}, a “prompt” involves providing natural language instructions or commands to guide an AI model in task completion. This approach offers several benefits, including increased model effectiveness, reduced training time and costs, and improved interpretability and controllability.
Notable works in this area include those by \cite{jiang-etal-2020-know,liu23acm,brown2020nips}.

\subsection{Tabular Data Synthesis}
Existing approaches for tabular data synthesis can be categorized into four main groups:

\noindent
\textbf{Probabilistic Models.} 
These models leverage probabilistic techniques to synthesize data. 
For instance, Gaussian copula models \cite{synthetic_data_vault} are effective for continuous variables but not for categorical ones. 
Conversely, Bayesian networks \cite{bayesian_networks_1,bayesian_networks_2} are adept at handling categorical data but struggle with continuous variables.
% \textbf{Probabilistic Model Based:} For example, Gaussian copula models continuous variables but not categorical~\cite{synthetic_data_vault}. Bayesian networks handle categorical but not continuous data~\cite{bayesian_networks_1,bayesian_networks_2}. 

\noindent
\textbf{Generative Adversarial Networks.} 
Generative Adversarial Networks (GANs) have been widely used to generate tabular data. 
MedGAN \cite{MedGAN} and RGAN \cite{rgan} produce healthcare records but face challenges with mixed data types. 
TableGAN \cite{tablegan} employs Convolutional Neural Networks (CNNs), demonstrating that synthetic data can perform comparably to real data. 
CTGAN and TVAE \cite{pr3-ctgan} address multimodality with column-specific preprocessing and Variational Gaussian Mixture (VGM) models. 
Other notable works include \cite{ctgan,Corrgan,PATE-gan,erhgan}.
% \textbf{GAN Based:} GAN methods generate tabular data~\cite{GANs}. MedGAN~\cite{MedGAN} and RGAN~\cite{rgan} produce healthcare records but struggle with mixed data types. TableGAN~\cite{tablegan} uses CNNs, showing synthetic data performs similarly to real data. CTGAN and TVAE~\cite{pr3-ctgan} handle multimodality with column-specific preprocessing and VGM. Other works include \cite{ctgan,Corrgan,PATE-gan,erhgan}.

\noindent
\textbf{Diffusion Models.}
These approaches utilize diffusion models for data synthesis. TabDDPM \cite{akim2023tabddpm} models both categorical and continuous values but encounters difficulties with correlations. 
SOS \cite{kim2022sos} uses Score-based Generative Models (SGMs) to handle imbalanced data, though it lacks the ability to condition on both data types.
% \textbf{Diffusion Model Based:} TabDDPM~\cite{akim2023tabddpm} and SOS~\cite{kim2022sos} use diffusion models. TabDDPM models categorical and continuous values but struggles with correlations. SOS uses SGMs for imbalanced data. These lack the ability to condition on both data types.

\noindent
\textbf{Language Models.} 
Self-attention models, which have revolutionized NLP \cite{vaswani2017attention}, have also been adapted for tabular data synthesis. 
These include encoding models \cite{lan2020albert,devlin2018bert}, sequence-to-sequence models \cite{seq2seq1}, and auto-regressive models \cite{auto-gre}. 
Transformers have been applied to table classification \cite{ft-transformer} and joint table-text representations \cite{under_rpt,table-to-text-tablegpt}. 
GReaT \cite{borisov2023great} and TapTap \cite{zhang2023taptap} generate synthetic tables but encounter limitations with wide tables.

\section{The Task of Tabular Data Synthesis }
% \textbf{Table Generation.} 
To find a data synthesizer $G$ learnt from a table $D$ and using $G$ to generate
a synthetic table $D_{syn}$. The objective of table generation is to produce data that is similar in distribution to the original data. We evaluate the generator $G$ from multiple perspectives: (1) Machine Learning Efficiency: When we train a classifier or regressor on the generated dataset, can it achieve the accuracy achieved by training on the original dataset? (2) Data Augmentation: When we add synthetic data to the original data, will it enhance the classification/regression task? (3) The difference between the synthetic data and the original data. We hope that the synthetic data is not a copy of the original data.

\section{Methods}
\label{method}
This section introduces the AIGT method for generating tabular data using prompt-based enhancement. 
% 这里增加表格生成的定义
% This section begins with an introduction to the definition of table generation and then introduces the AIGT method for generating tabular data using prompt-based enhancement. 
As illustrated in Figure~\ref{Fig: overview}, AIGT comprises five main stages:
(1) \textbf{Prompt Design:} 
Construct a prompt based on the table’s caption information and column names.
%We construct a prompt based on the table's caption information and table column name information.
(2) \textbf{Textual Encoding:} 
Convert table features and their values into sentences, concatenate these into prompts, and construct data suitable for model input.
%This process involves converting table features and their values into sentences, concatenating these into prompts, and constructing data that can be fed into models.
% Before starting training AIGT, we collected and cleaned 1000 public datasets from the OpenML website. 
(3) \textbf{Training Procedure:} 
Utilize a pre-trained Large Language Model (LLM) on an extensive corpus, and then conduct specific fine-tuning for downstream tables.
%Use a pre-trained Large Language Model (LLM) in the pre-trained corpus. 
% (4) \textbf{Fine-tuning:} 
% Conduct specific fine-tuning for downstream tables.
%Corresponding fine-tuning training for specific downstream tables. 
(4) \textbf{Generation:} 
Generate samples using the fine-tuned auto-regressive language model.
%Generating samples using the fine-tuned auto-regressive language model. 
To support tables of any size, unrestricted by the number of features, we proposed a \textbf{Partitioning Algorithm for Long Tokens} to enhance the scalability of our model.

% \subsection{~\ours{} Pre-training Corpus}
\subsection{Prompt Design} \label{prompt_dg}
To enhance the understanding of table structures and semantics in language models, we utilize the metadata of tables to construct prompts. In our framework,  the metadata of the table comprises information from two parts.
\begin{itemize}
     \item Caption information: the description of the dataset, including the purpose of the dataset, background information.
     \item Feature information: the names of features, the target column for prediction, and the meanings associated with the features, especially the meanings of certain abbreviations.
 \end{itemize} 

We define the function $\textbf{prompt}$ that can process metadata, which is implemented by calling the GPT3.5\footnote{OpenAI API: gpt-3.5-turbo}. The corresponding code is available in the appendix \ref{ap:ptp}.

% The provided Prompt is 
%  "Following is a description of a dataset, a profile of an object from the dataset, and a target description. The objective is to predict the target based on the information provided about the object. Dataset description: \{table caption\}, feature name: \{columns\}, Target: \{target column\},
%  the meaning of columns:\{the meaning of features\}.
%  Information to be returned includes: 1) a brief summary of the table description (such as field and background); 2) the target columns; 3) the features and their explanations. Here is an example output: The dataset is about economics, the target is income,  the features along with their explanations are as follows: ID represents a unique identifier for each user; Age denotes the age of each user. Make it brief but informative. Try to limit it to 200 words."

% The specific implementation methods and examples are shown in the Appendix \ref{ap:ptp}.

% By integrating these elements and domain-specific knowledge into the prompts, the model can gain a comprehensive understanding of the table's structure and semantics.

\subsection{Textual Encoding}

\noindent
\textbf{Feature Serialization.}
The standard large language model expects text as input.
Thus,
~\ours{} transforms each row from the dataset into a text format.
% must convert each row of the dataset into a textual representation.
We follow the previous work \cite{borisov2023great} by serializing each sample into a text sequence. 
By concatenating the feature names and values of the table into sentences, that is, "[Feature] is [Value]". 
Considering that tabular data follows the invariance of feature arrangement, we apply an arrangement function $P$ to randomly shuffle the order of features when encoding a table. 
Formally, given a table $D = \{(x_i, y_i)\}$, 
Let $x_{ij}$ be the $j$-th feature value of the i-th sample and $F_{i}$ represents the feature name in the $i$-th column. 
The textual encoding is to transform the i-th sample $x_{i}$ into a splice of sentences separated by commas $T_{i} = (T_{i,k_{1}} ,T_{i,k_{2}}, ..., T_{i,k_{m}})$, where $[k_{1}, k_{2}, ..., k_{m}] = P([1, 2, ..., m])$ and $T_{i,j} = (F_{j}\ is\ x_{ij})$ and $m$ is the total number of columns.

% \subsection{~\ours{} Training Strategy}

\begin{figure}[t]
  \centering
  \includegraphics[width=0.9\linewidth]{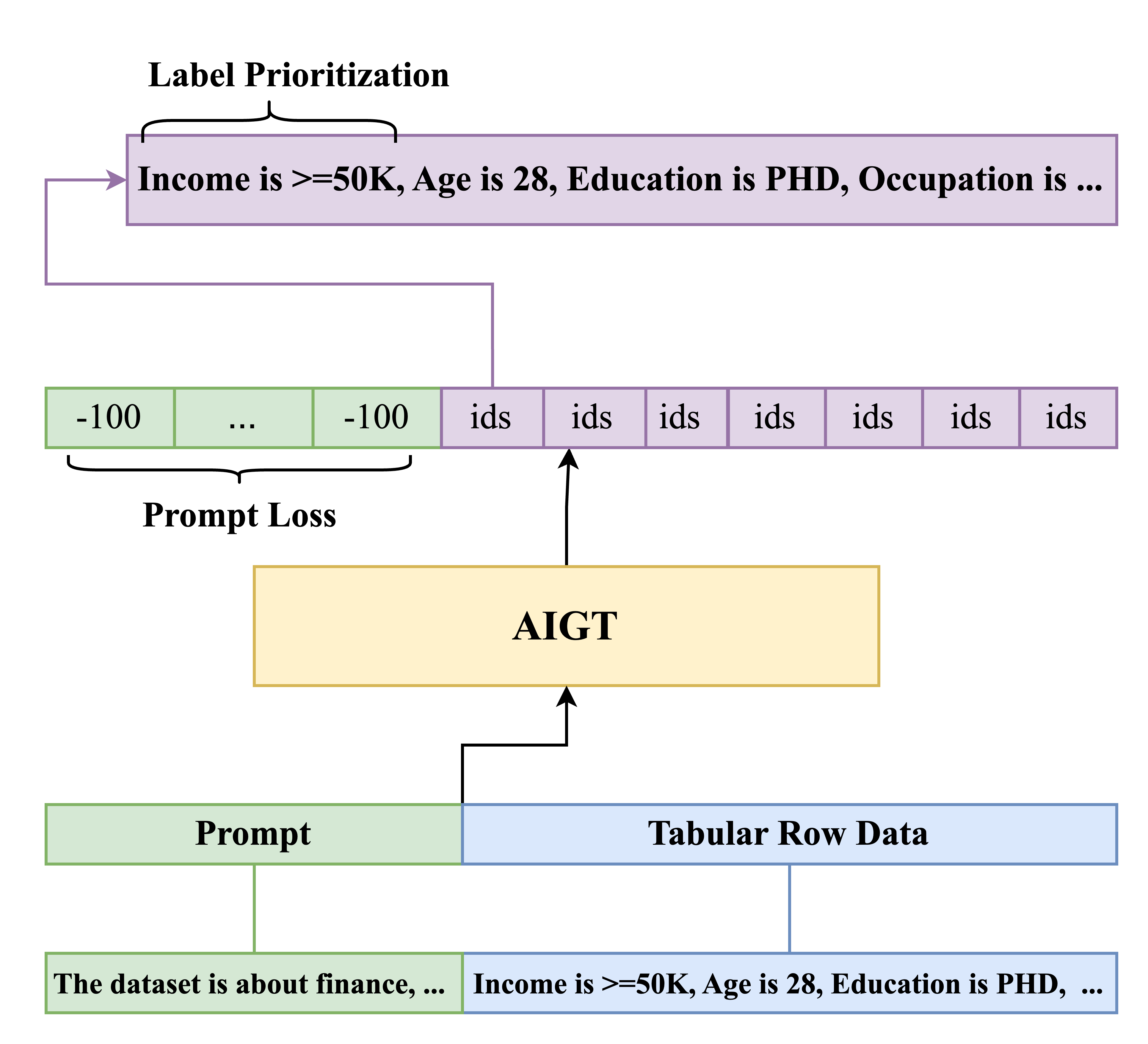}
  \caption{Training Strategies for ~\ours{}. Here, Prompt's losses are not calculated, the label feature was fixed in the first place, the other features were pre-mutated randomly.}
  \label{Fig: train}
% \vspace{-0.5cm} %这里图片与正文空两行正文的距离
\end{figure}

\noindent
\textbf{Label Prioritization.} 
We believe that the meaning of features and labels in tables is different, and labels are a summary of all features. 
It is crucial to prioritize the label column during the serialization process. 
By placing the label column at the forefront, we ensure that the model can immediately recognize the target variable, which can enhance the overall understanding and performance of the model. 
Therefore, given a table $D = \{(x_i, y_i)\}$ along with a prompt, which is generated based on the table metadata (refer to Section \ref{prompt_dg}), where $F_{y}$ represents the name of the label, the encoded sentence becomes $T_{i} = (\textbf{prompt} ~ F_{y}\ is\ y_{i}, T_{i,k_{1}} ,T_{i,k_{2}}, ..., T_{i,k_{m}})$, where $[k_{1}, k_{2}, ..., k_{m}] = P([1, 2, ..., m])$.

\noindent
\textbf{Prompt-Enhanced Loss.} 
Similar to the loss design of most prompt engineering methods ~\cite{prompt}, ~\ours{}'s training strategy only calculates non prompt loss calculations and ignores the loss calculations of the prompt itself, to achieve better table generation quality. 
We show these strategies in Figure~\ref{Fig: train}.

\subsection{Training Procedure}
\noindent
\textbf{Pre-Training.} 
We perform pre-training on a large-scale dataset upstream. Specifically, following the first two steps, we convert each row of table data into text to form the pre-training corpus $\mathcal{T}$.
Each sentence $t\in\mathcal{T}$ can be encoded into a sequence of tokens using $tokenize(t)=(w_{1}, ..., w_{N})$. 
In general, ~\ours{} factorizes the probability of generating $t$ in an auto-regressive manner as $p(t) =\prod \limits_{i=1}^N p(w_{i}|w_{1}, ..., w_{i-1})$. 
During pre-training, ~\ours{} is optimized towards maximizing the probability $\prod \limits_{i=1}^{\left\| \mathcal{T}\right\|} p(i)$ on the entire pre-training corpus. 
The pre-training process can initiate with an auto-regressive language model, thereby capitalizing on the extensive knowledge
that these models have already acquired.
% The pre-training can start with an auto-regressive LM, so that ~\ours{} can benefit from the common knowledge already learned by these LMs. 

\noindent
\textbf{Fine-tuning.} 
The fine-tuning of AIGT on downstream tables follows a similar process as pre-training. 
The only difference is that fine-tuning aims to target specific downstream tables.

\subsection{Generation}  
\noindent
\textbf{Sampling.} 
We have trained an auto-regressive model $\textbf{q}$ through fine-tuning on the text training dataset. 
This model predicts the potential subsequent labels $w_{1},..., w_{k-1}$ for the classification output distribution $z=\textbf{q}(w_{1},... w_{k-1})$. 
Multiple sampling strategies can be utilized in this scenario. 
Typically, the next token $w$ is selected through weighted sampling from the output $z$ of the LLM, guided by a temperature parameter $T>0$,
\begin{equation}
\label{eq:sample}
    p(w|w_{1}, ..., w_{k-1}) = \frac{e^{(z_{w}/T)}}{\sum_{w'\in \mathcal{W}} e^{(z_{w'}/T)}}.
\end{equation}
% \vspace{0.1cm}

Following GReaT \cite{borisov2023great},
the model is initialized with specific conditions 
and LLM is tasked with sampling the remaining tokens to complete the feature vector in its textual representation. 
Although we fix the position of the label in the first place during training, it is possible to generate data according to a specific feature column; one simply needs to prepend the corresponding label. 
For example, the provided condition can be "\textbf{prompt}   $[Label]$ is $[Value]$".
Here, the prompt is generated from the semantic information of the table metadata as Section~\ref{prompt_dg}.

\noindent
\textbf{Re-Labeling.} 
We observe that in TapTap~\cite{zhang2023taptap}, the re-labeling through the table prediction model effectively enhances the ability to predict labels of synthesized data.  
Given the original train datasets $D = \{(x_i,y_i)\}$, similar to labeling technique, we fine-tune AIGT on it and  generate synthesized data $D' = \{(x'_i,y'_i)\}$. 
Then, a tabular predictor $P$, \emph{e.g.}, LightGBM, is trained to fit $D$, and then the synthesis label $y'_i$ can be replaced by $y'_i = P(x'_i)$.

% \subsection{~\ours{} Data Relabeling}
% We have observed that in TapTap~\cite{zhang2023taptap}, the re-labeling through the table prediction model effectively enhances the ability to predict labels of synthesized data.  Similar to the technique of labeling, given the original train datasets $D = (x_i,y_i)$, we fine-tune AIGT on it and  generate synthesized data $D' = (x'_i,y'_i)$. Then, A tabular predictor $P$(e.g., LightGBM) is trained to fit $D$, then the label of synthesis $y'_i$ can be replaced by $y'_i = P(x'_i)$. The ablation experiment in Section~\ref{abs_exp} indicates that the re-labeling strategy plays a key role in boosting the performance of our approach.

\subsection{Partitioning Algorithm for Long Tokens}
% In practical industrial scenarios, datasets are usually large in scale, with many features, and each feature name is also lengthy. This will result in the number of tokens for $T_{i}$ greatly exceeding the input limit supported by LLM, which is a huge challenge for LLM-based methods. Based on this, we propose a long token partitioning algorithm that can perfectly couple previous training and generation methods.

In practical industrial scenarios, datasets are large with numerous features, leading to the number of tokens to exceed LLM input limits.
To address this scalability challenge for LLM-based methods, we propose a long-token partitioning algorithm that integrates seamlessly with existing training and generation methods.

\begin{figure}[t]
  \centering
  \includegraphics[width=\linewidth]{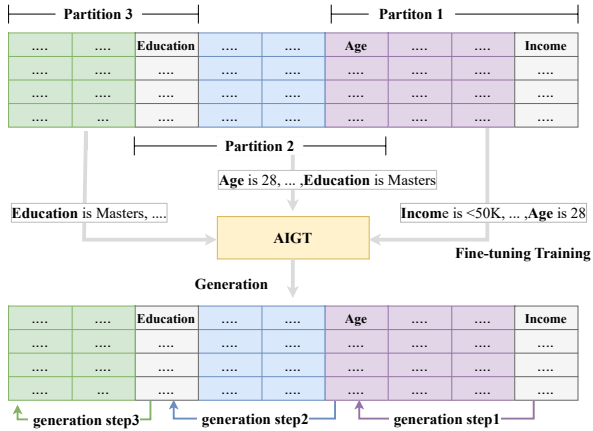}
  \caption{The Process of Long Token Partition Algorithm in~\ours{}. Divide the table into sub-tables based on columns, with overlaps between the columns of the sub-tables. Use the data from the sub-tables to train and generate the model.}
  \label{Fig:long-token-train}
\end{figure} 

\noindent
\textbf{Partition Training}. 
As shown in the Figure~\ref{Fig:long-token-train}, first, we partition the features to ensure that there are some overlapping columns between each region, and then perform mixed training to enable the large language model to learn the feature associations of different partitions. 
The function of the cover column is to alleviate to some extent the missing feature associations between different regions. 
In our industrial scenario experiment, the number of overlapping
columns is set to 1, and due to the small number of features on academic datasets, partitioning algorithms are not required.

\noindent
\textbf{Partition Generation}.
When generating, it is generated from the back to the front. 
After each partition is generated, the remaining partitions are generated using the cover part of the generated partition as the starting distribution column, and finally merge the partitions.

\section{Experiments} 
In Section \ref{exp:setup}, we outline the datasets, the baseline methods utilized in our experiments. 
In Section \ref{exp:all_exp}, we conducted extensive experiments, encompassing the machine learning efficiency of generated data, Distance to Closest Record (DCR) distance, data augmentation, and the application of our partitioning algorithm across both public and industrial datasets. 
To further show the excellence of our approach, we include additional analysis in the appendix \ref{ap:other_exp}, such as discriminative metrics and feature statistical similarity between generated and original data.

% In this section, we conducted extensive experiments on several benchmark datasets to evaluate the effectiveness and superiority of~\ours{}.

\subsection{Experimental Setup}
\label{exp:setup}
\paragraph{Datasets.}
The experimental dataset consists of three parts as following:

(1) \textbf{Upstream Large-scale Cross-table Pre-training Dataset.} We collected approximately 1000 tables with  their metadata, sourced primarily from OpenML \footnote{https://www.openml.org/},
    referred to as ~\datas{}. We have made ~\datas{} open-source in the hope of contributing to and advancing the community focused on pre-training tabular data. More information about ~\datas{} can be seen in the appendix \ref{ap:stab}.
    
(2) \textbf{Downstream Open Public Dataset.} For downstream tabular tasks, we utilize a diverse set of 20 public benchmark tabular datasets  to test the efficacy of our model.
These datasets, sourced from OpenML, UCI repository and Kaggle, contain both binary, multi-class classification and Regression tasks. 
Information  regarding  each dataset is provided in the Table ~\ref{tab:data_info} .
% To maintain a fair and unbiased evaluation, we have deliberately excluded these downstream tabular tasks from the pre-training corpus. Information  regarding  each dataset is provided in the Table ~\ref{tab:data_info} .
    
(3) \textbf{Ailpay Dataset.} Datasets in real-world industrial settings generally comprise a multitude of column names. To evaluate the efficiency and performance of long token partitioning, we utilized two industrial datasets from Alipay's risk control system. We show the details in  Table \ref{tab:alipay_data_info}.

\begin{table}[t]% h asks to places the floating element [h]ere.
  \centering
  \caption{Information of Downstream Tabular Datasets. Dataset abbreviations are in
parentheses. \# Num and \# Cat represent the number of continuous
and categorical columns.}
\resizebox{\linewidth}{!}{
  \begin{tabular}{llllll}
   \toprule
   Abbr& Name           & \#Samples & \# Num & \# Cat & Task type  \\  
   \midrule
       BW  & Breast-w           & 699       & 9     & 0     & Binclass  \\  
       CG  & Credit-g           & 1000      & 7     & 13    & Binclass   \\
       DI  & Diabetes           & 768       & 8     & 0     & Binclass  \\
       EL  & Electricity        & 45312     & 8     & 0     & Binclass \\
       SI  & Sick               & 3772      & 7     & 22    & Binclass  \\
       AD  & Adult              & 48842     & 15    & 6     & Binclass  \\
       WI  & Wilt               & 4839      & 6     & 5     & Binclass \\
       CM  & Cmc                & 1473      & 9     & 0     & Multiclass \\ 
       VH  & Vehicle            & 846       & 18    & 0     & Multiclass \\
       SA  & Satimage           & 6430      & 36    & 0     & Multiclass \\
       AF  & Analcatdata\_dmft  & 661       & 2     & 2     & Multiclass \\
       CR  & Car                & 1728      & 0     & 6     & Multiclass \\
       SE  & Segment            & 2310      & 15    & 0     & Multiclass \\
       EU  & Eucalyptus         & 736       & 14    & 5     & Multiclass  \\ 
       LO  & Loan               & 5000      & 12    & 0     & Binclass  \\
       HE  & Heloc              & 10460     & 22    & 1     & Binclass \\
       CA  & California Housing & 20640     & 8     & 0     & Regression \\
       AG  & Crab Age           & 3894      & 7     & 1     & Regression \\
       IN  & Insurance          & 1338      & 3     & 3     & Regression \\
       KI  & King               & 21613     & 18    & 1     & Regression \\ 
       \bottomrule
  \end{tabular}}
  \label{tab:data_info}
  % \vspace{1cm}
\end{table}

\begin{table}
  \centering
  \caption{Statistical Information of Alipay Datasets.~The two tables come from Alipay's financial risk control system, and their goals are binary, PR means the proportion of instances that are labeled as positive.} 
  \resizebox{\linewidth}{!}{
  \begin{tabular}{l|lllll}
       \toprule
       Dataset Name & \# columns & \# train & \# valid & \# test & PR \\  
       \midrule
       % SYH-xiaoheng & 251 & 484671 & 53853 & 8797 & 0.02 \\ 
       SYH & 251 & 466320 & 51814 & 29187 & 0.06 \\ 
       % FP  & 350 & 922208 & 184627 & 91285502 & 0.05 \\ 
       NonBD & 45 & 6148904 & 1537226 & 960766 & 0.05 \\ 
       \bottomrule
  \end{tabular}}
  \label{tab:alipay_data_info}
  \vspace{-0.3cm}
\end{table}

\paragraph{Baseline Methods.}
We evaluate ~\ours{} alongside five other SOTA tabular data synthesis algorithms: CTGAN, TVAE, TabDDPM, GReaT and TapTap.  CTGAN \cite{ctgan} based on generative adversarial networks \cite{GANs} for tabular data, allowing the generation process to be conditional only on a single discrete feature. The same author proposed TVAE \cite{ctgan}, a variational autoencoder (VAE) for tabular data. TabDDPM \cite{akim2023tabddpm} adapts diffusion models to tabular data.  GReaT and TapTap, which utilize language-based approaches, adopt a pre-trained DistilGPT-2
model. This DistilGPT-2 framework also serves as the foundational model for ~\ours{}. To illustrate the potential of larger models, our approach incorporates Llama3.1-8B as a foundational model, denoted as ~\ours{}-L.

% \section{Reproducibility Details}
\paragraph{Baseline Implementation.}
For CTGAN and TVAE,  we set the training epochs to 300, except for those datasets that have less than 5k data. Due to their small sample size, we will set a larger number of training epochs to 500, to ensure better training results on these small datasets. For the diffsion method TabDDPM, we employ default settings.
For GReaT, TapTap and ~\ours{}, we use the distilGPT-2 as framework. We pretrained AIGT for 10w steps with the learning rate $1\times 10^{-4}$ . We finetune the GReaT, TapTap and 
~\ours{} for 100 epoch. The batch size is 32 for all datasets.  We use the AdamW optimizer for the proposed generative models, with the learning rate $5\times 10^{-5}$. For ~\ours{}-L,  We use the AdamW optimizer with the learning rate $1\times 10^{-5}$.

\paragraph{Environment} Experiments run on a machine equipped with
4 NVIDIA A100-SXM4-80GB GPU and 100 GB RAM , Intel(R) Xeon(R) Platinum 8369B CPU @ 2.90GHz CPU under Ubuntu 20.04 with 64 cores.

\subsection{Overall Performance}
\label{exp:all_exp}
% In this section, we report the overall performance of ~\ours{}. The results are shown in Table~\ref{tab:lgb_mle}--\ref{tab:aigt_part}.

In this section, we will demonstrate the performance of the proposed AIGT method through multiple experiments. Additionally, we also present the effects of the partitioning algorithm on both public and industrial datasets.

\paragraph{Machine Learning Efficiency (MLE).}
In this section, we compare \ours{} to alternative generative models in terms of machine learning efficiency.
Each dataset was split into two parts: 80\% for training purposes and 20\% reserved for testing. 
Initially, each generative algorithm is trained on the training data. Subsequently, the trained model is utilized to generate synthetic data of equivalent size. This synthetic data is then used to train a classification/regression model, which is then evaluated using the real test set. 
We expect that for high-quality synthetic data, models trained on this data will perform comparably to those trained on real data. To assess the effectiveness of the machine learning models, we apply the LightGBM model, a leading GBDT method, to evaluate their efficiency. We adopt the AUC score as the evaluation metric for classification tasks and employ $R^2$ score for regression tasks. For
a fair comparison we use the the standard hyper-parameter tuning budget of 50 trials. Our full search space is provided in the Appendix \ref{ap:tune_param} and all the experimental results are averaged over 10 different random seeds. The result was showed in Table \ref{tab:lgb_mle},
Note that we match or exceed state-of-the-art on 14 out of 20 datasets.

\begin{table*}
    \centering
    \caption{ML efficiency experiment. We used 20 real-world datasets, for classification tasks, auc score is reported. For regression datasets, $R^2$ is reported. The values of machine learning efficiency computed with regards to the state-of-the-art tuned LightGBM model. The best results are marked in bold, while the second best results are marked with underscores.}
    \resizebox{\linewidth}{!}{
    \begin{tabular}{l l l l l l l l l l l}
    \toprule
    Dataset & BW & CM  & CG  & DI  & VH  & EL & SA & EU & SI & AF  \\
    \midrule

    \textbf{Real}   & 99.2\scalebox{0.75}{$\pm$0.1} 
                    & 74.9\scalebox{0.75}{$\pm$0.1}
                    & 76.9\scalebox{0.75}{$\pm$1.8}
                    & 82.7\scalebox{0.75}{$\pm$1.7}
                    & 93.9\scalebox{0.75}{$\pm$0.0}
                    & 97.6\scalebox{0.75}{$\pm$0.1}
                    & 99.1\scalebox{0.75}{$\pm$0.0}
                    & 89.1\scalebox{0.75}{$\pm$0.0}
                    & 96.3\scalebox{0.75}{$\pm$0.5}
                    & 52.5\scalebox{0.75}{$\pm$1.1}\\
                    
    % \hline
    \midrule
    \textbf{CTGAN}  & 97.5\scalebox{0.75}{$\pm$2.6} 
                    & 50.1\scalebox{0.75}{$\pm$3.1}
                    & 52.4\scalebox{0.75}{$\pm$4.0}
                    & 75.0\scalebox{0.75}{$\pm$2.6}
                    & 58.8\scalebox{0.75}{$\pm$2.4}
                    & 82.4\scalebox{0.75}{$\pm$0.4}
                    & 95.4\scalebox{0.75}{$\pm$0.2}
                    & 53.1\scalebox{0.75}{$\pm$3.5}
                    & 65.8\scalebox{0.75}{$\pm$0.8}
                    & \underline{53.8\scalebox{0.75}{$\pm$2.4}}\\
                    
    \textbf{TVAE}   & \underline{98.9\scalebox{0.75}{$\pm$0.3}}
                    & 59.3\scalebox{0.75}{$\pm$0.9}
                    & 74.8\scalebox{0.75}{$\pm$2.1}
                    & 80.3\scalebox{0.75}{$\pm$0.9}
                    & 86.6\scalebox{0.75}{$\pm$0.4}
                    & 84.0\scalebox{0.75}{$\pm$1.1}
                    & 97.3\scalebox{0.75}{$\pm$1.0}
                    & 84.9\scalebox{0.75}{$\pm$0.8}
                    & 93.2\scalebox{0.75}{$\pm$1.5}
                    & 53.0\scalebox{0.75}{$\pm$2.3}\\  
                    
    \textbf{TabDDPM}    &\textbf{99.0\scalebox{0.75}{$\pm$0.2}}
                        & \underline{71.7\scalebox{0.75}{$\pm$0.8}}
                        & 71.0\scalebox{0.75}{$\pm$1.2}
                        & 78.5\scalebox{0.75}{$\pm$2.2}
                        & 59.8\scalebox{0.75}{$\pm$0.7}
                        & 89.1\scalebox{0.75}{$\pm$0.1}
                        & 78.8\scalebox{0.75}{$\pm$4.2}
                        & 69.5\scalebox{0.75}{$\pm$0.7}
                        & 96.5\scalebox{0.75}{$\pm$0.7}
                        & 52.5\scalebox{0.75}{$\pm$0.6}\\

    \textbf{GReaT}  & 98.2\scalebox{0.75}{$\pm$0.4} 
                    & 56.4\scalebox{0.75}{$\pm$3.2}
                    & 53.9\scalebox{0.75}{$\pm$3.1}
                    & 59.2\scalebox{0.75}{$\pm$0.3}
                    & 64.3\scalebox{0.75}{$\pm$4.5}
                    & 92.3\scalebox{0.75}{$\pm$0.3}
                    & 95.9\scalebox{0.75}{$\pm$0.3}
                    & 81.1\scalebox{0.75}{$\pm$2.1}
                    & 96.2\scalebox{0.75}{$\pm$1.2}
                    & 53.1\scalebox{0.75}{$\pm$2.2}\\

    \textbf{TapTap}  & \textbf{99.0\scalebox{0.75}{$\pm$0.2}} 
                    & 71.5\scalebox{0.75}{$\pm$0.4}
                    & 76.3\scalebox{0.75}{$\pm$1.6}
                    & 79.7\scalebox{0.75}{$\pm$0.2}
                    & \underline{90.8\scalebox{0.75}{$\pm$1.6}}
                    & 91.1\scalebox{0.75}{$\pm$0.2}
                    & \underline{97.8\scalebox{0.75}{$\pm$0.2}}
                    & \underline{86.8\scalebox{0.75}{$\pm$0.7}}
                    & \textbf{98.4\scalebox{0.75}{$\pm$0.8}}
                    & 53.2\scalebox{0.75}{$\pm$0.7}\\
    
    \midrule
    \textbf{AIGT}   & 98.5\scalebox{0.75}{$\pm$0.1} 
                    & 71.6\scalebox{0.75}{$\pm$0.6}
                    & \textbf{78.1\scalebox{0.75}{$\pm$2.3}}
                    & \underline{81.7\scalebox{0.75}{$\pm$0.5}}
                    & \textbf{91.5\scalebox{0.75}{$\pm$1.1}}
                    & \textbf{94.3\scalebox{0.75}{$\pm$0.2}}
                    & \underline{97.8\scalebox{0.75}{$\pm$0.1}}
                    & \textbf{86.9\scalebox{0.75}{$\pm$1.0}}
                    & \underline{97.9\scalebox{0.75}{$\pm$1.6}}
                    & 53.8\scalebox{0.75}{$\pm$0.9}\\

    \textbf{AIGT-L}   & 98.2\scalebox{0.75}{$\pm$0.3}
                    & \textbf{72.1\scalebox{0.75}{$\pm$0.6}}
                    & \underline{76.5\scalebox{0.75}{$\pm$1.3}}
                    & \textbf{82.2\scalebox{0.75}{$\pm$0.1}}
                    & \textbf{91.5\scalebox{0.75}{$\pm$0.1}}
                    & \underline{93.3\scalebox{0.75}{$\pm$0.5}}
                    & \textbf{97.9\scalebox{0.75}{$\pm$0.4}}
                    & 86.6\scalebox{0.75}{$\pm$0.5}
                    & 97.8\scalebox{0.75}{$\pm$1.4}
                    & \textbf{54.4\scalebox{0.75}{$\pm$1.0}}\\
    
    \bottomrule
    \end{tabular}
    }
    \resizebox{\linewidth}{!}{
    \begin{tabular}{l l l l l l l l l l l}
    \toprule
    Dataset & AD & WI  & CR  & SE  & LO  & HE & IN & AG & CA & KI  \\
    \midrule
    \textbf{Real}   & 92.7\scalebox{0.75}{$\pm$1.3} 
                    & 99.1\scalebox{0.75}{$\pm$0.0}
                    & 100.0\scalebox{0.75}{$\pm$0.0}
                    & 99.3\scalebox{0.75}{$\pm$0.0}
                    & 99.8\scalebox{0.75}{$\pm$0.0}
                    & 80.9\scalebox{0.75}{$\pm$0.1}
                    & 88.0\scalebox{0.75}{$\pm$0.0}
                    & 53.2\scalebox{0.75}{$\pm$0.3}
                    & 85.3\scalebox{0.75}{$\pm$0.2}
                    & 87.8\scalebox{0.75}{$\pm$0.1}\\
    \midrule
    \textbf{CTGAN}  & 89.4\scalebox{0.75}{$\pm$0.2} 
                    & 53.2\scalebox{0.75}{$\pm$3.6}
                    & 56.7\scalebox{0.75}{$\pm$1.9}
                    & 86.6\scalebox{0.75}{$\pm$2.0}
                    & 49.0\scalebox{0.75}{$\pm$1.4}
                    & 40.6\scalebox{0.75}{$\pm$2.0}
                    & 37.7\scalebox{0.75}{$\pm$6.4}
                    & 27.6\scalebox{0.75}{$\pm$1.6}
                    & 50.1\scalebox{0.75}{$\pm$0.1}
                    & 55.3\scalebox{0.75}{$\pm$1.8}\\
                    
    \textbf{TVAE}   & 87.4\scalebox{0.75}{$\pm$0.4} 
                    & 80.6\scalebox{0.75}{$\pm$1.8}
                    & 89.1\scalebox{0.75}{$\pm$2.3}
                    & 97.7\scalebox{0.75}{$\pm$0.2}
                    & 97.2\scalebox{0.75}{$\pm$0.4}
                    & 61.0\scalebox{0.75}{$\pm$4.0}
                    & 61.0\scalebox{0.75}{$\pm$3.2}
                    & 20.5\scalebox{0.75}{$\pm$1.0}
                    & 69.0\scalebox{0.75}{$\pm$0.3}
                    & 72.3\scalebox{0.75}{$\pm$2.5}\\ 
                    
    \textbf{TabDDPM}    & \underline{90.6\scalebox{0.75}{$\pm$0.1}} 
                        & \textbf{98.9\scalebox{0.75}{$\pm$0.3}}
                        & 99.3\scalebox{0.75}{$\pm$0.2}
                        & 98.3\scalebox{0.75}{$\pm$0.1}
                        & 98.6\scalebox{0.75}{$\pm$0.2}
                        & 74.6\scalebox{0.75}{$\pm$0.5}
                        & 81.0\scalebox{0.75}{$\pm$0.1}
                        & 45.8\scalebox{0.75}{$\pm$0.3}
                        & 80.0\scalebox{0.75}{$\pm$0.2}
                        & 75.0\scalebox{0.75}{$\pm$3.0}\\

    \textbf{GReaT}  & \textbf{90.8\scalebox{0.75}{$\pm$0.3}}
                & 96.1\scalebox{0.75}{$\pm$1.1} 
                & 91.4\scalebox{0.75}{$\pm$2.0} 
                & 98.1\scalebox{0.75}{$\pm$2.0} 
                & 97.8\scalebox{0.75}{$\pm$0.2} 
                & 75.3\scalebox{0.75}{$\pm$0.4} 
                & 57.9\scalebox{0.75}{$\pm$5.1} 
                & 41.2\scalebox{0.75}{$\pm$0.8} 
                & 75.6\scalebox{0.75}{$\pm$0.3} 
                & 61.8\scalebox{0.75}{$\pm$2.1} \\

    \textbf{TapTap} & 90.0\scalebox{0.75}{$\pm$0.4} 
                & 97.7\scalebox{0.75}{$\pm$0.9} 
                & 99.4\scalebox{0.75}{$\pm$0.2} 
                & \textbf{98.6\scalebox{0.75}{$\pm$0.1}}
                & \textbf{99.5\scalebox{0.75}{$\pm$0.1}} 
                & \underline{79.8\scalebox{0.75}{$\pm$0.3}}
                & \textbf{88.0\scalebox{0.75}{$\pm$0.1}}
                & 51.4\scalebox{0.75}{$\pm$0.3} 
                & \underline{82.6\scalebox{0.75}{$\pm$0.1}} 
                & \underline{87.7\scalebox{0.75}{$\pm$0.3}} \\
    
    \midrule
    \textbf{AIGT}   & 89.8\scalebox{0.75}{$\pm$0.2} 
                    & 97.4\scalebox{0.75}{$\pm$0.8} 
                    & \underline{99.8\scalebox{0.75}{$\pm$0.1}} 
                    & \textbf{98.6\scalebox{0.75}{$\pm$0.1}} 
                    & \underline{99.4\scalebox{0.75}{$\pm$0.2}}
                    & \textbf{80.3\scalebox{0.75}{$\pm$0.1}} 
                    & \underline{87.9\scalebox{0.75}{$\pm$0.2}} 
                    & \textbf{53.3\scalebox{0.75}{$\pm$0.2}} 
                    & 82.4\scalebox{0.75}{$\pm$0.1}
                    & \textbf{88.0\scalebox{0.75}{$\pm$0.2}}\\
                    
    \textbf{AIGT-L}  & 86.7\scalebox{0.75}{$\pm$1.4} 
                    & \underline{98.8\scalebox{0.75}{$\pm$0.2}}
                    & \textbf{99.9\scalebox{0.75}{$\pm$0.7}}
                    & \underline{98.4\scalebox{0.75}{$\pm$0.1}}
                    & \underline{99.4\scalebox{0.75}{$\pm$0.2}}
                    & 79.0\scalebox{0.75}{$\pm$0.3}
                    & 87.4\scalebox{0.75}{$\pm$0.5}
                    & \underline{53.1\scalebox{0.75}{$\pm$0.5}}
                    & \textbf{83.2\scalebox{0.75}{$\pm$0.1}}
                    & 87.5\scalebox{0.75}{$\pm$0.4}\\
    \bottomrule
    \end{tabular}
    }
    \label{tab:lgb_mle}
    % \vspace{0.5cm}
    \end{table*}

% dcr的画图
% \begin{figure}[t]
%     \centering
%     \subfloat[CA]{
%         \includegraphics[width=\linewidth]{figures/ca_dcr.png}}\\
%     \subfloat[SA]{
%         \includegraphics[width=\linewidth]{figures/sa_dcr.png}}
%     \caption{Distance to closest record (DCR) distribution.“Original” denotes the
% DCR of the original test set with respect to the original train set. The experimental results illustrate that each
% method does not copy samples from the train set.}
%     \label{Fig: dcr}
% \end{figure}

\begin{figure*}[t]
    \includegraphics[width=\linewidth]{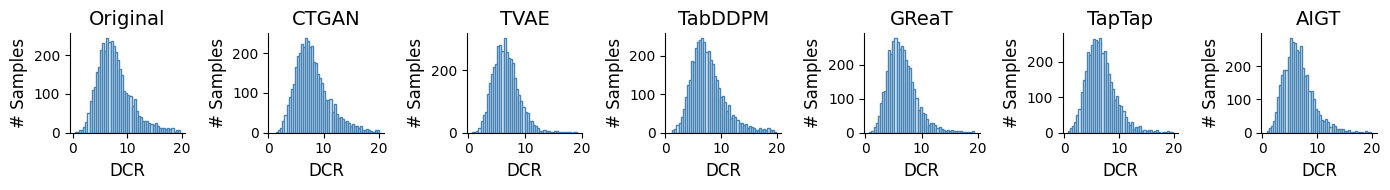}
    \caption{Distance to closest record (DCR) distribution for the California Housing dataset. “Original” denotes the
DCR of the original test set with respect to the original train set. The experimental results illustrate that each
method does not copy samples from the train set.}
\label{Fig: dcr}
\end{figure*}

\paragraph{Distance to closest records  histogram} 
To verify that the generated data is similar to the original sample, rather than an exact replica, this metric calculates the distance from the nearest record in the original training 
dataset $D_{train}$. For each synthesized record $s$, it is given by DCR ($s$)=$\min\{distance(s, s_i)|s_i\in D_{train} \}$.
As a distance measure, we use the L1 norm for numerical features. For categorical features, we set the difference to 0, otherwise it is set to 1. 
% We compare the distribution of the minimal distances of the generated samples to the training data
% set. 
% The visualization of the minimum distance distribution can be found in Figure \ref{Fig: dcr}.
Note that models such as CTGAN and TabDDPM have a fixed DCR score. However, generative algorithms based on language models can produce more novel samples by adjusting the temperature coefficient. For the sampling step of GReaT, TapTap and ~\ours{}, we set the temperature parameter $T$ to 0.7 for all datasets. 
We compare the distribution of the minimal distances of the generated samples to the training data
set. 
The visualization of the minimum distance distribution can be found in Figure \ref{Fig: dcr}.

% % dcr的画图
% \begin{figure}[ht]
%     \centering
%     \subfloat[CA]{
%         \includegraphics[width=\linewidth]{figures/ca_dcr.png}}\\
%     \subfloat[SA]{
%         \includegraphics[width=\linewidth]{figures/sa_dcr.png}}
%     \caption{Distance to closest record (DCR) distribution.“Original” denotes the
% DCR of the original test set with respect to the original train set. The experimental results illustrate that each
% method does not copy samples from the train set.}
%     \label{Fig: dcr}
% \end{figure}

\paragraph{Data Augmentation.} When dealing with datasets that are relatively small in sample size, our approach is to boost the performance of machine learning models by supplementing these datasets with synthetic data. We integrate the synthetic data with the original training data to create an augmented training set. Following this, we conduct model training on this enlarged training set and assess its performance on the test set. The results are presented in Table \ref{tab:add_aug}, which show that \ours{} is able to perform better than all baseline methods on most datasets.

\paragraph{Effectiveness of Partition Generation Algorithm.}
To validate our partitioning algorithm, we selected datasets with more than 20 columns from public datasets and  two Real-world industrial Ailpay datasets. We divided the academic datasets and the NonBD table into two partitions, and partitioned the SYH table into five segments. Utilizing the synthetic data generated through our partitioning algorithm, we evaluated the ML efficiency  of the synthetic datasets and compared it with the generation methods that do not employ language models. The results of the partitioning algorithm are shown in Table \ref{tab:aigt_part}. It can be seen that even with the partitioning algorithm, our method still outperforms models not using language models in efficiency.

% % dcr的画图
% \begin{figure}[t]
%     \centering
%     \subfloat[CA]{
%         \includegraphics[width=\linewidth]{figures/ca_dcr.png}}\\
%     \subfloat[SA]{
%         \includegraphics[width=\linewidth]{figures/sa_dcr.png}}
%     \caption{Distance to closest record (DCR) distribution.“Original” denotes the
% DCR of the original test set with respect to the original train set. The experimental results illustrate that each
% method does not copy samples from the train set.}
%     \label{Fig: dcr}
% \end{figure}

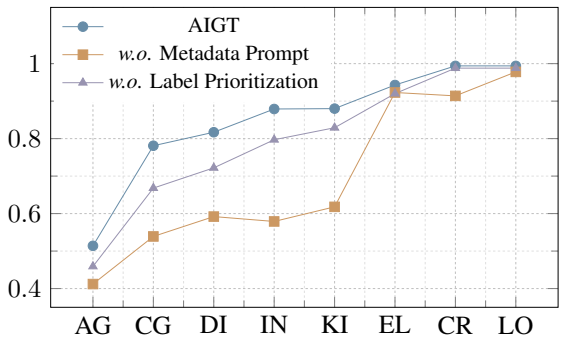
\begin{figure}[ht]
\begin{tikzpicture}[scale=0.9]
\begin{axis}[
    width=9cm, % 设置宽度
    height=6cm, % 设置高度，根据长宽比16:9计算
    symbolic x coords={AG,CG,DI,IN,KI,EL,CR,LO},  
    xtick=data,     
     ymin=0.35, ymax=1.15,
     ytick={0.4,0.6,0.8,1.0},
     grid=both, % 启用主要和次要网格线
    major grid style={line width=.1pt, draw=gray!50,  dash pattern=on 1pt off 1pt}, % 主要网格线样式
    minor grid style={line width=.1pt, draw=gray!25, dash pattern=on 1pt off 1pt}, % 次要网格线样式
    minor tick num=1, % 次要刻度数量（在两个主要刻度之间）
     legend style={at={(0.28,1)},
    anchor=north,legend columns=1,font=\small,draw=none,},
]
\addplot[color=mycolor3!80!black, mark=*] coordinates {
(AG,0.514)
(CG,0.781)
(DI,0.817)
(IN,0.879)
(KI,0.880)
(EL,0.943)
(CR,0.994)
(LO,0.994)
% (IN,0.879)
% (AG,0.514)
% (KI,0.880)
};
\addlegendentry{AIGT}
\addplot[color=mycolor2!80!black, 
mark=square*] coordinates {
(AG,0.412)
(CG,0.539)
(DI,0.592)
(IN,0.579)
(KI,0.618)
(EL,0.923)
(CR,0.914)
(LO,0.978)
% (IN,0.579)
% (AG,0.412)
% (KI,0.618)
};
\addlegendentry{\textit{w.o.} Metadata Prompt}
\addplot[color=mycolor1!80!black, mark=triangle*] coordinates {
(AG,0.459)
(CG,0.668)
(DI,0.722)
(IN,0.797)
(KI,0.829)
(EL,0.920)
(CR,0.988)
(LO,0.988)
% (IN,0.797)
% (AG,0.459)
% (KI,0.829)
};
\addlegendentry{\textit{w.o.} Label Prioritization}
\end{axis}
\end{tikzpicture}
 \caption{Ablation experiments related to training strategies. The y-axis is the average metric values across all
datasets  with LightGBM.}
\label{Fig: ab-all}
\vspace{-0.3cm}
\end{figure}

% \paragraph{Data Augmentation.} When dealing with datasets that are relatively small in sample size, our approach is to boost the performance of machine learning models by supplementing these datasets with synthetic data. We integrate the synthetic data with the original training data to create an augmented training set. Following this, we conduct model training on this enlarged training set and assess its performance on the test set. The results are presented in Table \ref{tab:add_aug}, which show that \ours{} is able to perform better than all baseline methods on most datasets.

% \paragraph{Effectiveness of Partition Generation Algorithm.}
% To validate our partitioning algorithm, we selected datasets with more than 20 feature columns from public datasets, as well as two Real-world industrial Ailpay datasets. We divided the academic datasets and the NonBD table into two partitions, and partitioned the SYH table into five segments. Utilizing the synthetic data generated through our partitioning algorithm, we evaluated the ML efficiency  of the synthetic datasets and compared it with the generation methods that do not employ language models. The results of the partitioning algorithm are shown in Table \ref{tab:aigt_part}. It can be seen that even with the partitioning algorithm, our method still outperforms models not using language models in efficiency.

\begin{table*}
    \centering
    \caption{Data Augmentation. "Real" means training with the original data. Each generative methods means training with the original data plus the synthetic data.}
    \resizebox{\linewidth}{!}{
    \begin{tabular}{l l l l l l l l l l l}
    \toprule
    Dataset & BW & CM  & CG  & DI  & VH  & EL & SA & EU & SI & AF  \\
    \midrule

    \textbf{Real}   & 99.2\scalebox{0.75}{$\pm$0.1} 
                    & 74.9\scalebox{0.75}{$\pm$0.1}
                    & 76.9\scalebox{0.75}{$\pm$1.8}
                    & 82.7\scalebox{0.75}{$\pm$1.7}
                    & 93.9\scalebox{0.75}{$\pm$0.0}
                    & 97.6\scalebox{0.75}{$\pm$0.1}
                    & 99.1\scalebox{0.75}{$\pm$0.0}
                    & 89.1\scalebox{0.75}{$\pm$0.0}
                    & 96.3\scalebox{0.75}{$\pm$0.5}
                    & 52.5\scalebox{0.75}{$\pm$1.1}\\
                    
    % \hline
    \midrule
    \textbf{CTGAN}  & 99.0\scalebox{0.75}{$\pm$0.1} 
                    & 72.8\scalebox{0.75}{$\pm$0.7}
                    & 72.6\scalebox{0.75}{$\pm$0.4}
                    & 80.2\scalebox{0.75}{$\pm$0.7}
                    & 92.4\scalebox{0.75}{$\pm$0.0}
                    & 95.5\scalebox{0.75}{$\pm$0.1}
                    & 99.0\scalebox{0.75}{$\pm$0.0}
                    & 88.1\scalebox{0.75}{$\pm$0.0}
                    & 99.8\scalebox{0.75}{$\pm$0.5}
                    & 53.3\scalebox{0.75}{$\pm$1.1}\\
                    
    \textbf{TVAE}   & 99.1\scalebox{0.75}{$\pm$0.1} 
                    & 73.2\scalebox{0.75}{$\pm$0.7}
                    & 75.6\scalebox{0.75}{$\pm$1.6}
                    & \textbf{83.0\scalebox{0.75}{$\pm$0.7}}
                    & 92.0\scalebox{0.75}{$\pm$0.6}
                    & 95.4\scalebox{0.75}{$\pm$0.1}
                    & 99.0\scalebox{0.75}{$\pm$0.0}
                    & 89.1\scalebox{0.75}{$\pm$0.8}
                    & 99.7\scalebox{0.75}{$\pm$0.1}
                    & 53.1\scalebox{0.75}{$\pm$1.1}\\  
                    
    \textbf{TabDDPM}    & 99.1\scalebox{0.75}{$\pm$0.2} 
                        & 73.3\scalebox{0.75}{$\pm$1.6}
                        & 74.7\scalebox{0.75}{$\pm$1.2}
                        & 82.2\scalebox{0.75}{$\pm$1.2}
                        & 92.5\scalebox{0.75}{$\pm$0.0}
                        & 95.5\scalebox{0.75}{$\pm$0.1}
                        & \textbf{99.1\scalebox{0.75}{$\pm$0.0}}
                        & 89.1\scalebox{0.75}{$\pm$0.5}
                        & 99.7\scalebox{0.75}{$\pm$0.1}
                        & 53.0\scalebox{0.75}{$\pm$0.5}\\

    \textbf{GReaT}  & 99.1\scalebox{0.75}{$\pm$0.2} 
                    & 71.8\scalebox{0.75}{$\pm$0.8}
                    & 72.6\scalebox{0.75}{$\pm$1.9}
                    & 79.5\scalebox{0.75}{$\pm$2.0}
                    & 91.9\scalebox{0.75}{$\pm$0.3}
                    & 96.8\scalebox{0.75}{$\pm$0.0}
                    & 98.6\scalebox{0.75}{$\pm$0.1}
                    & 88.8\scalebox{0.75}{$\pm$0.7}
                    & 99.8\scalebox{0.75}{$\pm$0.0}
                    & 53.1\scalebox{0.75}{$\pm$0.2}\\

    \textbf{TapTap}  & \textbf{99.2\scalebox{0.75}{$\pm$0.2}} 
                    & \textbf{73.8\scalebox{0.75}{$\pm$0.4}}
                    & 75.7\scalebox{0.75}{$\pm$1.4}
                    & 80.9\scalebox{0.75}{$\pm$1.0}
                    & 93.2\scalebox{0.75}{$\pm$0.6}
                    & 97.1\scalebox{0.75}{$\pm$0.2}
                    & \textbf{99.1\scalebox{0.75}{$\pm$0.4}}
                    & \textbf{89.2\scalebox{0.75}{$\pm$2.0}}
                    & 99.8\scalebox{0.75}{$\pm$2.4}
                    & 53.1\scalebox{0.75}{$\pm$0.8}\\
    
    \midrule
    \textbf{AIGT}   & \textbf{99.2\scalebox{0.75}{$\pm$1.1}} 
                    & 73.5\scalebox{0.75}{$\pm$0.9}
                    & \textbf{79.5\scalebox{0.75}{$\pm$2.6}}
                    & 81.7\scalebox{0.75}{$\pm$1.7}
                    & \textbf{93.6\scalebox{0.75}{$\pm$0.4}}
                    & \textbf{97.9\scalebox{0.75}{$\pm$0.2}}
                    & 99.0\scalebox{0.75}{$\pm$0.6}
                    & \textbf{89.2\scalebox{0.75}{$\pm$2.0}}
                    & \textbf{99.9\scalebox{0.75}{$\pm$0.2}}
                    & \textbf{53.8\scalebox{0.75}{$\pm$0.5}}\\
     \textbf{AIGT-L}   & \textbf{99.2\scalebox{0.75}{$\pm$0.1}} 
                    & 74.2\scalebox{0.75}{$\pm$0.6}
                    & 76.6\scalebox{0.75}{$\pm$0.4}
                    & 82.3\scalebox{0.75}{$\pm$0.1}
                    & 93.0\scalebox{0.75}{$\pm$0.4}
                    & 97.1\scalebox{0.75}{$\pm$0.1}
                    & 99.0\scalebox{0.75}{$\pm$0.6}
                    & 88.6\scalebox{0.75}{$\pm$0.4}
                    & \textbf{99.9\scalebox{0.75}{$\pm$0.2}}
                    & \textbf{53.8\scalebox{0.75}{$\pm$1.0}}\\
    \bottomrule
    \end{tabular}
    }
    \resizebox{\linewidth}{!}{
    \begin{tabular}{l l l l l l l l l l l}
    \toprule
    Dataset & AD & WI  & CR  & SE  & LO  & HE & IN & AG & CA & KI  \\
    \midrule
    \textbf{Real}   & 92.7\scalebox{0.75}{$\pm$1.3} 
                    & 99.1\scalebox{0.75}{$\pm$0.0}
                    & 100.0\scalebox{0.75}{$\pm$0.0}
                    & 99.3\scalebox{0.75}{$\pm$0.0}
                    & 99.8\scalebox{0.75}{$\pm$0.0}
                    & 80.9\scalebox{0.75}{$\pm$0.1}
                    & 88.0\scalebox{0.75}{$\pm$0.0}
                    & 53.2\scalebox{0.75}{$\pm$0.3}
                    & 85.3\scalebox{0.75}{$\pm$0.2}
                    & 87.8\scalebox{0.75}{$\pm$0.1}\\
    \midrule
    \textbf{CTGAN}  & 92.4\scalebox{0.75}{$\pm$0.1} 
                    & 97.5\scalebox{0.75}{$\pm$0.6}
                    & 96.4\scalebox{0.75}{$\pm$1.1}
                    & 99.3\scalebox{0.75}{$\pm$0.1}
                    & 98.9\scalebox{0.75}{$\pm$0.4}
                    & 75.4\scalebox{0.75}{$\pm$0.6}
                    & 59.3\scalebox{0.75}{$\pm$6.0}
                    & 49.8\scalebox{0.75}{$\pm$0.6}
                    & 81.4\scalebox{0.75}{$\pm$0.5}
                    & 82.3\scalebox{0.75}{$\pm$1.1}\\
                    
    \textbf{TVAE}   & 92.4\scalebox{0.75}{$\pm$0.1} 
                    & 97.2\scalebox{0.75}{$\pm$0.8}
                    & 98.0\scalebox{0.75}{$\pm$0.6}
                    & 99.3\scalebox{0.75}{$\pm$0.1}
                    & \textbf{99.8\scalebox{0.75}{$\pm$0.5}}
                    & 77.3\scalebox{0.75}{$\pm$0.3}
                    & 82.4\scalebox{0.75}{$\pm$3.2}
                    & 49.3\scalebox{0.75}{$\pm$0.4}
                    & 81.5\scalebox{0.75}{$\pm$0.1}
                    & 82.4\scalebox{0.75}{$\pm$0.3}\\ 
                    
    \textbf{TabDDPM}    & 92.3\scalebox{0.75}{$\pm$0.1} 
                        & \textbf{99.5\scalebox{0.75}{$\pm$0.2}}
                        & 99.8\scalebox{0.75}{$\pm$0.1}
                        & 99.4\scalebox{0.75}{$\pm$0.0}
                        & \textbf{99.8\scalebox{0.75}{$\pm$0.1}}
                        & 80.2\scalebox{0.75}{$\pm$0.3}
                        & 86.5\scalebox{0.75}{$\pm$0.6}
                        & 53.0\scalebox{0.75}{$\pm$0.5}
                        & 84.1\scalebox{0.75}{$\pm$0.2}
                        & 82.4\scalebox{0.75}{$\pm$1.0}\\

    \textbf{GReaT}  & \textbf{92.7\scalebox{0.75}{$\pm$0.1}} 
                & 99.1\scalebox{0.75}{$\pm$0.3} 
                & 99.8\scalebox{0.75}{$\pm$0.2} 
                & 99.3\scalebox{0.75}{$\pm$0.1} 
                & \textbf{99.8\scalebox{0.75}{$\pm$0.1}} 
                & 80.4\scalebox{0.75}{$\pm$0.1} 
                & 81.9\scalebox{0.75}{$\pm$4.3} 
                & 53.0\scalebox{0.75}{$\pm$0.5} 
                & 82.3\scalebox{0.75}{$\pm$0.2} 
                & 78.0\scalebox{0.75}{$\pm$1.8} \\

    \textbf{TapTap}  & 91.9\scalebox{0.75}{$\pm$0.2} 
                    & 98.9\scalebox{0.75}{$\pm$0.7} 
                    & 99.9\scalebox{0.75}{$\pm$0.6} 
                    & \textbf{99.8\scalebox{0.75}{$\pm$0.0}} 
                    & \textbf{99.8\scalebox{0.75}{$\pm$0.4}} 
                    & 81.0\scalebox{0.75}{$\pm$0.4} 
                    & 88.1\scalebox{0.75}{$\pm$0.2} 
                    & 53.0\scalebox{0.75}{$\pm$0.6} 
                    & \textbf{85.2\scalebox{0.75}{$\pm$2.0}} 
                    & 87.6\scalebox{0.75}{$\pm$0.5} \\
    
    \midrule
    \textbf{AIGT}   & 91.7\scalebox{0.75}{$\pm$0.4} 
                    & 99.3\scalebox{0.75}{$\pm$0.2} 
                    & \textbf{100.0\scalebox{0.75}{$\pm$0.0}} 
                    & 99.7\scalebox{0.75}{$\pm$0.8} 
                    & 99.7\scalebox{0.75}{$\pm$0.4} 
                    & \textbf{81.1\scalebox{0.75}{$\pm$0.1}} 
                    & \textbf{88.2\scalebox{0.75}{$\pm$0.1}} 
                    & \textbf{54.2\scalebox{0.75}{$\pm$0.2}} 
                    & \textbf{85.2\scalebox{0.75}{$\pm$0.1}} 
                    & 87.3\scalebox{0.75}{$\pm$3.0} \\
    \textbf{AIGT-L}   & \textbf{92.2\scalebox{0.75}{$\pm$0.7}} 
                    & 99.3\scalebox{0.75}{$\pm$0.4}
                    & \textbf{100.0\scalebox{0.75}{$\pm$0.4}}
                    & 99.3\scalebox{0.75}{$\pm$0.4}
                    & \textbf{99.8\scalebox{0.75}{$\pm$0.4}}
                    & 79.2\scalebox{0.75}{$\pm$0.2}
                    & 88.1\scalebox{0.75}{$\pm$0.8}
                    & \textbf{54.2\scalebox{0.75}{$\pm$0.1}}
                    & \textbf{85.2\scalebox{0.75}{$\pm$0.1}}
                    & \textbf{88.7\scalebox{0.75}{$\pm$1.6}}\\
    \bottomrule
    \end{tabular}
    }
    \label{tab:add_aug}
    % \vspace{0.5cm}
    \end{table*}

\begin{table*}[!ht]
    \centering
    \caption{ML efficiency experiment for partition generation algorithm. The public datasets and the NonBD table were devided into two partitions, and partitioned the SYH table
    into five segments. Below the backbone model is
LightGBM. \textcolor{red}{\ding{55}} indicates that the calculation cannot be performed due to the excessive columns in the table.}
    % \resizebox{\linewidth}{!}{
    \scalebox{0.85}{
    \begin{tabular}{l l l l l l l l l}
    \toprule
    Dataset  & CG  &  SA & EU & SI & HE &KI &SYH &NonBD  \\
    \midrule
    \textbf{CTGAN}  & 52.4\scalebox{0.75}{$\pm$4.0} 
                    & 95.4\scalebox{0.75}{$\pm$0.2}
                    & 53.1\scalebox{0.75}{$\pm$3.5}
                    & 65.8\scalebox{0.75}{$\pm$0.8}
                    & 40.6\scalebox{0.75}{$\pm$2.0}
                    & 55.3\scalebox{0.75}{$\pm$1.8}
                    & 45.7\scalebox{0.75}{$\pm$0.8}
                    & 78.6\scalebox{0.75}{$\pm$0.6}
                \\
                 
    \textbf{TVAE}   & 74.8\scalebox{0.75}{$\pm$2.1} 
                    & 97.3\scalebox{0.75}{$\pm$1.0}
                    & 84.9\scalebox{0.75}{$\pm$0.9}
                    & 93.2\scalebox{0.75}{$\pm$1.5}
                    & 61.0\scalebox{0.75}{$\pm$4.0}
                    & 72.3\scalebox{0.75}{$\pm$2.5}
                    & 54.6\scalebox{0.75}{$\pm$0.5}
                    & 86.3\scalebox{0.75}{$\pm$1.2}
                    \\
                 
    \textbf{TabDDPM}    & 71.0\scalebox{0.75}{$\pm$1.2} 
                        & 78.8\scalebox{0.75}{$\pm$4.2}
                        & 69.5\scalebox{0.75}{$\pm$0.7}
                        & 96.5\scalebox{0.75}{$\pm$0.7}
                        & 74.6\scalebox{0.75}{$\pm$0.5}
                        & 75.0\scalebox{0.75}{$\pm$3.0}
                        & 56.1\scalebox{0.75}{$\pm$0.7}
                        & 79.6\scalebox{0.75}{$\pm$1.8}
                        \\

    \textbf{AIGT}  & 78.1\scalebox{0.75}{$\pm$2.3} 
                    & 97.8\scalebox{0.75}{$\pm$0.1}
                    & 86.9\scalebox{0.75}{$\pm$1.0}
                    & 97.9\scalebox{0.75}{$\pm$1.6}
                    & 80.3\scalebox{0.75}{$\pm$0.1}
                    & 88.0\scalebox{0.75}{$\pm$0.2}
                    & \quad\textcolor{red}{\ding{55}}
                    & \quad\textcolor{red}{\ding{55}}
                 \\
    
    \hline 
    \textbf{AIGT-part}   & 74.3\scalebox{0.75}{$\pm$1.1} 
                    & 97.8\scalebox{0.75}{$\pm$0.2}
                    & 83.9\scalebox{0.75}{$\pm$0.8}
                    & 96.4\scalebox{0.75}{$\pm$0.4}
                    & 78.6\scalebox{0.75}{$\pm$0.1}
                    & 86.6\scalebox{0.75}{$\pm$0.8}
                    & 57.2\scalebox{0.75}{$\pm$1.6}
                    & 88.2\scalebox{0.75}{$\pm$0.2}
                   \\
    \bottomrule
    \end{tabular}}
    % }

    \label{tab:aigt_part}
    \end{table*}
    
% \subsection{Ablation Experiment}
\subsection{Ablation Analysis}
\label{abs_exp}

\paragraph{Effects Analysis of Training Strategy.}  
We selected 8 datasets and tested ablation experiments
under the following different conditions: (1) \textbf{\textit{w.o.} Metadata Prompt}. This refers to  \ours{} without the metadata from tables serving as fixed prompts. (2) \textbf{\textit{w.o.} Label Prioritization}. We consider labels as features, shuffling them with other features for use in the training and generation processes, rather than moving the label column to the first position during the serialization process.
 The experimental results are shown in Figure \ref{Fig: ab-all}, we can see that the prompt-enhanced method indeed has a good gain, demonstrating the effectiveness of our method.

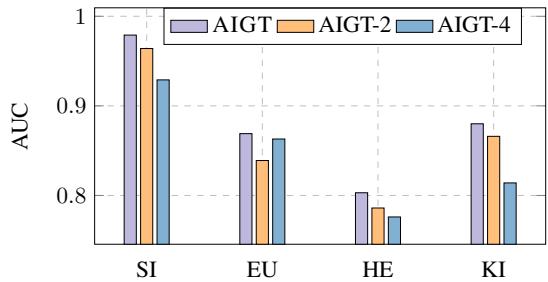
\begin{figure}[h]
\begin{tikzpicture}[scale=0.8]
\begin{axis}[
    width=9cm,
    height=5.5cm,
    ymajorgrids,
      xmajorgrids,
      grid style=dashed,
    ybar,
    bar width=0.2cm,
    enlargelimits=0.15,
    legend style={at={(0.55,1)},
    anchor=north,legend columns=-1},
    ylabel={AUC},
    x tick style={opacity=0},
    symbolic x coords={SI, EU, HE, KI},
    xtick=data,
    nodes near coords style={invisible},
    ]
\addplot[fill=mycolor1, area legend] coordinates {(SI,0.979) (EU,0.869) (HE,0.803) (KI,0.880)};
\addlegendentry{\ours{}}
\addplot[fill=mycolor2, area legend] coordinates {(SI,0.964) (EU,0.839) (HE,0.786) (KI,0.866)};
\addlegendentry{AIGT-2}
\addplot[fill=mycolor3, area legend] coordinates {(SI,0.929) (EU,0.863) (HE,0.776) (KI,0.814)};
\addlegendentry{AIGT-4}
\end{axis}
\end{tikzpicture}
\caption{Analysis of the number of partitions for partition
generation algorithm. "AIGT" signifies no partitioning, while "AIGT-2" denotes partitioning into two sections,"AIGT-4" denotes partitioning into four sections.}
\label{Fig: part_abs}
\end{figure}

\paragraph{Performance Analysis of Partition Algorithm.}
We evaluate the impact of the number of partitions in our partitioning algorithm on the synthesis of data. Four public datasets are selected for this purpose. The results are shown in Figure \ref{Fig: part_abs}. 
\section{Conclusion}
In this paper, we introduce a novel data-synthesis method for language models called ~\ours{}, enhanced with prompts. This method utilizes the metadata of tables, guiding the language model to generate data more effectively. 

Existing table generation methods based on language models are unable to tackle the issue of long tokens. 
To overcome this limitation, we've  designed a long-token
partitioning algorithm, can support the generation of tabular data at any scale. Our approach is more flexible, capable of handling tabular data with a larger number of feature columns.

\section*{Acknowledgments}
This work is supported by the Fundamental Research Funds for the Central Universities (226-2024-00049), the NSFC Grants (No.62206247) and the Pioneer R\&D Program of Zhejiang (No.2024C01035). The authors from Ant Group are supported by the Leading Innovative and Entrepreneur Team Introduction Program of Hangzhou (Grant No.TD2022005).

\section*{Limitations}
A primary limitation of our method is its processing speed. While we leverage the powerful capabilities of LLMs, this also results in increased running time and GPU memory consumption compared to more lightweight approaches such as GANs. Detailed comparisons are provided in appendix \ref{ap:time}. Furthermore, our current method for handling numerical values treats numbers as characters, applying tokenization and transformation without additional processing. This approach is unable to capture the magnitude relationships of numerical values, focusing solely on semantic similarity during tokenization. Future work will aim to develop more advanced encoding techniques for numerical values to address this limitation.

\section*{Ethics Statement}
In this paper, the ~\datas{} dataset we constructed is derived from publicly available datasets that have been openly shared on well-known machine learning platforms such as OpenML, ensuring that no private information is included. For the risk control data from Alipay that we used, since it is internal company data and has not been anonymized, we only present the algorithm's performance without disclosing the dataset to protect user privacy and comply with data security laws and regulations.
Additionally, our use of the GPT-3.5 API is conducted through compliant intermediaries and utilized by the enterprise in accordance with regulatory requirements. 
% We strictly adhere to all relevant ethical and legal standards to ensure that the use and handling of data meet the highest privacy and security standards.

% \newpage
\bibliography{custom}

\appendix

\section{Dataset}
\label{sec:appendix}

\subsection{More Information about ~\datas{}}
\label{ap:stab}
In this paper, we collected and filtered out 978
publicly available tabular datasets to construct the
pre-training corpus for ~\ours{}. Significant effort has been invested in conducting thorough data filtering and cleaning procedures to uphold dataset quality.
For each table, our data cleaning protocols include, but are not limited to:

\noindent
(1) \textit{Check the semantic degree of the column names.} For example, the column names \{\textit{user\_age}, weight, \textit{monthly\_income}\} have high semantic information, while the column names \{\textit{f1}, \textit{f2}, \textit{xyz}\} have low. We calculate a cumulative semantic relevance score for each table and, as part of our protocol, exclude tables which less than 50\% of the semantic score.

\noindent
(2) \textit{Check the missing values.} The datasets with more than 40\% missing values are discarded. Because too many missing values can easily lead to biased or inaccurate results in the pre-training phase.

\noindent
(3) \textit{Data Preprocessing.} For categorical features in the tables, we restore them to their original textual values whenever possible. As for numerical features, we employ normalization to mitigate the impact of inconsistent measurement units across different tables (e.g., kilograms vs. grams).

\subsection{More Details about Downstream Dataset}
\label{ap:ddd}
We provide the  urls of the public datasets in Table
\ref{ap:data_info}. 

\begin{table}[ht]% h asks to places the floating element [h]ere.
  \centering
  \caption{The urls of Downstream Tabular Datasets. }
\resizebox{\linewidth}{!}{
  \begin{tabular}{cl}
   \toprule
   Abbr  & Link\\  
   \midrule
       BW   &  https://www.openml.org/d/15 \\  
       CG    &  https://www.openml.org/d/45058 \\
       DI  &  https://www.openml.org/d/42608 \\
       EL   &  https://www.openml.org/d/151 \\
       SI    &  https://www.openml.org/d/41946 \\
       AD   &  https://www.openml.org/d/1590 \\
       WI   &  https://www.openml.org/d/40983\\
       CM   &  https://www.openml.org/d/45054\\ 
       VH   &  https://www.openml.org/d/42863\\
       SA   &  https://www.openml.org/d/42858\\
       AF   &  https://www.openml.org/d/469\\
       CR   &  https://www.openml.org/d/40975\\
       SE   &  https://www.openml.org/d/42860\\
       EU    &  https://www.openml.org/d/43925\\ 
       LO    &  https://www.kaggle.com/datasets/burak3ergun/loan-data-set\\
       HE   &  https://www.kaggle.com/datasets/averkiyoliabev/home-equity-line-of-creditheloc\\
       CA   &  https://www.kaggle.com/datasets/camnugent/california-housing-prices\\
       AG  &  https://www.kaggle.com/datasets/sidhus/crab-age-prediction\\
       IN   &  https://www.kaggle.com/datasets/mirichoi0218/insurance\\
       KI   &  https://www.kaggle.com/harlfoxem/housesalesprediction\\ 
       \bottomrule
  \end{tabular}}
  \label{ap:data_info}

\end{table}

% \begin{table}
% % \renewcommand\arraystretch{0.8}
%   \centering
%   \caption{Statistical Information of Alipay Datasets.~The two tables come from Alipay's financial risk control system, and their goals are binary, PR means the proportion of instances that are labeled as positive.} 
%   \resizebox{\linewidth}{!}{
%   \begin{tabular}{c|ccccc}
%        \toprule
%        Dataset Name & \# columns & \# train & \# valid & \# test & PR \\  
%        \midrule
%        % SYH-xiaoheng & 251 & 484671 & 53853 & 8797 & 0.02 \\ 
%        SYH & 251 & 466320 & 51814 & 29187 & 0.06 \\ 
%        % FP  & 350 & 922208 & 184627 & 91285502 & 0.05 \\ 
%        NonBD & 45 & 6148904 & 1537226 & 960766 & 0.05 \\ 
%        \bottomrule
%   \end{tabular}}
%   \label{ap:alipay_data_info}
%   \vspace{-0.3cm}
% \end{table}

\section{Prompt Templates}
\label{ap:ptp}
In our implementation, we use the gpt-3.5-turbo model via the OpenAI-API to construct  the function $\textbf{prompt}$. Here we presented the code to call the GPT API in Listing \ref{ls:call_gpt}:

\lstset{ %
  language=Python,                % 代码语言
  basicstyle=\ttfamily\footnotesize, % 代码字体和大小
  keywordstyle=\color{blue},      % 关键字颜色
  commentstyle=\color{gray},      % 注释颜色
  stringstyle=\color{darkgreen},        % 字符串颜色
  numbers=left,                   % 行号位置
  numberstyle=\tiny\color{gray},  % 行号样式
  stepnumber=1,                   % 行号步长
  numbersep=5pt,                  % 行号与代码间距
  backgroundcolor=\color{white},  % 代码背景颜色
  showspaces=false,               % 显示空格
  showstringspaces=false,         % 显示字符串中的空格
  showtabs=false,                 % 显示制表符
  frame=single,                   % 代码框
  rulecolor=\color{black},        % 框颜色
  tabsize=2,                      % 制表符大小
  captionpos=b,                   % 标题位置
  breaklines=true,                % 自动换行
  breakatwhitespace=false,        % 仅在空格处换行
  title=\lstname,                 % 显示文件名
  escapeinside={\%*}{*)},         % 在代码中使用LaTeX命令
  morekeywords={*,...}            % 添加更多关键字
}

% \begin{lstlisting}[language=Python, caption=Code for calling GPT3.5]
% import requests
% api_key = 'openai_api_key'
% url = 'https://api.openai.com/v1/engines/davinci-codex/completions'
% headers = {
%     'Content-Type': 'application/json',
%     'Authorization': f'Bearer {api_key}'
% }
% data = {
%     'prompt': '',
%     'max_tokens': 200
% }
% response = requests.post(url, headers=headers, json=data)
% if response.status_code == 200:
%     result = response.json()
%     print(result['choices'][0]['text'].strip())
% else:
%     print(f"Error: {response.status_code}")
%     print(response.json())
% \end{lstlisting}

 Our prompt input to gpt-3.5 is shown as Listing \ref{ls:pmp}. We utilize the table's metadata $\textbf{M}$ to construct prompt input. In our framework, we actively collect metadata from two sources:

 \begin{itemize}
     \item The caption information ($\textbf{C}$): the description of the dataset, including  the purpose of the dataset, background information.
     \item Feature information($\textbf{F}$): the name of columns and the meaning of columns.
 \end{itemize} 
 % Our prompt input to gpt-3.5 is shown as \ref{ls:pmp}.
 \begin{figure*}

 \begin{lstlisting}[language=Python, caption=Code for calling GPT3.5,label=ls:call_gpt]
import requests
api_key = 'openai_api_key'
url = 'https://api.openai.com/v1/engines/davinci-codex/completions'
headers = {
    'Content-Type': 'application/json',
    'Authorization': f'Bearer {api_key}'
}
data = {
    'prompt': '',
    'max_tokens': 200
}
response = requests.post(url, headers=headers, json=data)
if response.status_code == 200:
    result = response.json()
    print(result['choices'][0]['text'].strip())
else:
    print(f"Error: {response.status_code}")
    print(response.json())
\end{lstlisting}
\begin{lstlisting}[language=Python, caption=prompt construction,label=ls:pmp]
prompt=
"Following is a description of a dataset, a profile of an object from the dataset, and a target description. The objective is to predict the target based on the information provided about the object.\n 
Dataset description: \{table caption\},\n
feature name: \{columns\}.\n
Target: \{target column\}.\n
the meaning of columns:\{the meaning of features\}
Information to be returned includes: 1) a brief summary of the table description (such as field and background); 2) the target columns; 3) the features and their explanations. Here is an example output: The dataset is about economics, the target is income, and the features along with their explanations are as follows: ID represents a unique identifier for each user; Age denotes the age of each user. Make it brief but informative. Try to limit it to 200 words."
\end{lstlisting}
\end{figure*}

\section{Further Experiments}
\label{ap:other_exp}
% 只保留 其他两个指标

\subsection{Statistical similarity}
\begin{figure}[h]
    \centering
    \begin{tikzpicture}
    \small{
    \begin{axis}[
      ymajorgrids,
      xmajorgrids,
      grid style=dashed,
      xbar,
      height=.25\textwidth,
      width=.4\textwidth,
      bar width=.9em,
      enlarge y limits=0.2,
      symbolic y coords={{TabDDPM},{CTGAN},{TapTap},{TVAE},{AIGT}},
      ytick distance=1,
      y tick style={opacity=0},
      bar shift=0pt,
      enlarge x limits=0.3,xticklabel style={/pgf/number format/fixed,/pgf/number format/fixed zerofill,/pgf/number format/precision=1},
      y dir=reverse,
      nodes near coords,
      nodes near coords align={horizontal}]
      \addplot[fill=mycolor3!90, draw=mycolor3, text=black] coordinates {
         (3.79,{TabDDPM})
          (2.62,{CTGAN})
          (1.92,{TapTap})
          (1.80,{TVAE})
          (1.33,{AIGT})
          
    };
    \end{axis}
  }
    \end{tikzpicture}

    \caption{Correlation distance mean value for 20 datasets.}
    \label{Fig: corr_mean}
  \end{figure}
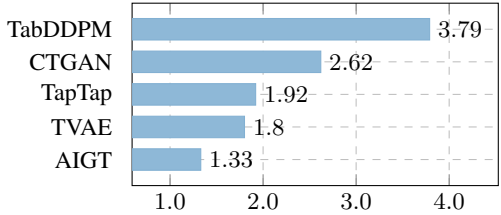

To accurately assess the dependencies among columns in synthetic data, we calculate pair-wise correlation matrices separately for both real and synthetic datasets. We use the Pearson correlation coefficient to analyze continuous variables, which produces results within the range of [-1, +1]. For categorical features, we use the uncertainty coefficient for evaluation, yielding values within the range of [0, 1]. Additionally, we use the correlation ratio to investigate the relationship between categorical and continuous variables, which also yields values within [0, 1]. Subsequently, we compute the Frobenius norm between the pairwise correlation matrices of the real and synthetic datasets and refer to it as the Correlation Distance. It is worth noting that a lower Correlation Distance value signifies a higher quality of data synthesis. The mean Correlation distance of 20 datasets was presented in figure \ref{Fig: corr_mean}.

\begin{table*}[!ht]
\caption{Discriminator measure.  A lower accuracy rate suggests a difficulty for the discriminator in discerning between artificial records and original samples. A completely indistinguishable dataset would yield an accuracy rate of 0.5. The superior results are highlighted in bold, while the results of the next best are underscored.  Best results are bold, second-best results are underlined. Results are averages over ten trials with different random seeds.}
\renewcommand\arraystretch{0.8}
\resizebox{\linewidth}{!}{
\scalebox{0.3}{
\tiny
\begin{tabular}{lllllll}
  \toprule
  Dataset   & \textbf{CTGAN} & \textbf{TVAE} &\textbf{TabDDPM} & \textbf{GReaT} & \textbf{TapTap}     & \textbf{AIGT}\\
  \midrule
    BW    & 99.9\scalebox{0.75}{$\pm$0.2}             & 99.8\scalebox{0.75}{$\pm$0.3}
          & \textbf{57.9\scalebox{0.75}{$\pm$2.8}}
          & 69.7\scalebox{0.75}{$\pm$1.1} & \underline{67.2\scalebox{0.75}{$\pm$2.6}}
          & 71.9\scalebox{0.75}{$\pm$2.5}\\
    
    CM    & 78.6\scalebox{0.75}{$\pm$2.3} & 89.9\scalebox{0.75}{$\pm$0.9} 
          & \underline{55.7\scalebox{0.75}{$\pm$1.4}}
          & 95.5\scalebox{0.75}{$\pm$0.4} & 95.7\scalebox{0.75}{$\pm$0.5} 
          & \textbf{54.0\scalebox{0.75}{$\pm$2.4}} \\
    
    CG     & \underline{90.0\scalebox{0.75}{$\pm$1.4}} & 99.6\scalebox{0.75}{$\pm$0.2}
            & \underline{75.3\scalebox{0.75}{$\pm$5.8}}
           & 99.3\scalebox{0.75}{$\pm$0.6} & 99.5\scalebox{0.75}{$\pm$0.5} 
           & \textbf{75.0\scalebox{0.75}{$\pm$4.5}}\\
    
    DI     & 97.7\scalebox{0.75}{$\pm$1.2} & 98.9\scalebox{0.75}{$\pm$0.8} 
            & 88.6\scalebox{0.75}{$\pm$2.3}
           & 76.1\scalebox{0.75}{$\pm$2.7} & \textbf{59.8\scalebox{0.75}{$\pm$3.5}} 
           & \underline{73.5\scalebox{0.75}{$\pm$2.7}}\\
 
    VH     & 99.4\scalebox{0.75}{$\pm$0.6} & 86.9\scalebox{0.75}{$\pm$2.9} 
           & 100.0\scalebox{0.75}{$\pm$0.0}
           & 88.4\scalebox{0.75}{$\pm$1.0} & \underline{78.8\scalebox{0.75}{$\pm$1.7}} 
           & \textbf{76.4\scalebox{0.75}{$\pm$3.0}}\\

   EL     & 99.7\scalebox{0.75}{$\pm$0.0} & 99.5\scalebox{0.75}{$\pm$0.1} 
           & 83.1\scalebox{0.75}{$\pm$0.4}
           & \textbf{68.7\scalebox{0.75}{$\pm$0.4}} & 100.0\scalebox{0.75}{$\pm$0.0} 
           & \underline{69.9\scalebox{0.75}{$\pm$0.3}}\\

   SA     & 99.9\scalebox{0.75}{$\pm$0.1} & 100.0\scalebox{0.75}{$\pm$0.0}
          & 100.0\scalebox{0.75}{$\pm$0.0}
          & \underline{77.5\scalebox{0.75}{$\pm$0.1}} & 100.0\scalebox{0.75}{$\pm$0.0} 
          & \textbf{78.7\scalebox{0.75}{$\pm$0.8}}\\
    
    EU      & \underline{99.8\scalebox{0.75}{$\pm$0.4}} & 99.9\scalebox{0.75}{$\pm$0.2} 
            & 100.0\scalebox{0.75}{$\pm$0.0}
            & 100.0\scalebox{0.75}{$\pm$0.0} & 100.0\scalebox{0.75}{$\pm$0.0}
            & \textbf{95.6\scalebox{0.75}{$\pm$0.2}} \\
    
    SI     & 98.7\scalebox{0.75}{$\pm$0.4} & 99.3\scalebox{0.75}{$\pm$0.3} 
           & 91.9\scalebox{0.75}{$\pm$0.9}
           & 73.8\scalebox{0.75}{$\pm$0.6} & \textbf{69.1\scalebox{0.75}{$\pm$0.1}} 
           & \underline{72.8\scalebox{0.75}{$\pm$0.7}}\\

    AF      & 67.9\scalebox{0.75}{$\pm$3.0} & 86.0\scalebox{0.75}{$\pm$2.2} 
            & \underline{53.0\scalebox{0.75}{$\pm$1.5}}
            & 71.4\scalebox{0.75}{$\pm$3.1} & 68.2\scalebox{0.75}{$\pm$4.0}
            & \textbf{51.4\scalebox{0.75}{$\pm$2.5}} \\

    AD      & 100.0\scalebox{0.75}{$\pm$0.0} & 100.0\scalebox{0.75}{$\pm$0.0}
            & \textbf{71.5\scalebox{0.75}{$\pm$0.4}}
            & 99.9\scalebox{0.75}{$\pm$0.0}  & 99.9\scalebox{0.75}{$\pm$0.1}
            & \underline{99.4\scalebox{0.75}{$\pm$0.0}}\\

    WI  & 84.1\scalebox{0.75}{$\pm$1.0} & 71.1\scalebox{0.75}{$\pm$1.4} 
        & \textbf{51.8\scalebox{0.75}{$\pm$0.2}}
        & 70.0\scalebox{0.75}{$\pm$0.4}& \underline{61.3\scalebox{0.75}{$\pm$0.7}} 
        & 69.8\scalebox{0.75}{$\pm$0.1} \\
    
    CR  & 69.2\scalebox{0.75}{$\pm$1.7} & 59.7\scalebox{0.75}{$\pm$1.5} 
        & 53.6\scalebox{0.75}{$\pm$1.8}
        & \underline{50.9\scalebox{0.75}{$\pm$0.2}} & 52.6\scalebox{0.75}{$\pm$0.5} 
        & \textbf{50.7\scalebox{0.75}{$\pm$0.2}}\\

    SE  & 100.0\scalebox{0.75}{$\pm$0.1} & 100.0\scalebox{0.75}{$\pm$0.1} 
        & 91.9\scalebox{0.75}{$\pm$0.6}
        & \underline{81.3\scalebox{0.75}{$\pm$0.2}} & 96.0\scalebox{0.75}{$\pm$0.1} 
        & \textbf{72.2\scalebox{0.75}{$\pm$0.1}} \\
    
    IN  & 89.7\scalebox{0.75}{$\pm$1.3} & 80.6\scalebox{0.75}{$\pm$1.1} 
        & \underline{67.9\scalebox{0.75}{$\pm$0.8}}
        & 82.6\scalebox{0.75}{$\pm$0.2} & 80.3\scalebox{0.75}{$\pm$0.3}
        & \textbf{67.8\scalebox{0.75}{$\pm$0.3}}\\

    HE & 98.5\scalebox{0.75}{$\pm$0.2} & \underline{96.1\scalebox{0.75}{$\pm$0.2}}
        & 85.7\scalebox{0.75}{$\pm$0.6}
       & 96.9\scalebox{0.75}{$\pm$0.4} & 96.6\scalebox{0.75}{$\pm$0.4} 
       & \textbf{78.8\scalebox{0.75}{$\pm$0.6}}\\

    CR  & 99.3\scalebox{0.75}{$\pm$0.2} & 98.6\scalebox{0.75}{$\pm$0.4}
        & \underline{87.8\scalebox{0.75}{$\pm$0.9}}
        & 99.2\scalebox{0.75}{$\pm$0.3} & 99.2\scalebox{0.75}{$\pm$0.4} 
        & \textbf{59.2\scalebox{0.75}{$\pm$0.7}} \\

    CA  & 91.5\scalebox{0.75}{$\pm$0.3}  & 88.7\scalebox{0.75}{$\pm$0.4} 
        & \textbf{65.0\scalebox{0.75}{$\pm$0.1}}
        & 76.5\scalebox{0.75}{$\pm$0.6} & 77.5\scalebox{0.75}{$\pm$0.6}
        & \underline{75.4\scalebox{0.75}{$\pm$0.6}}  \\
    
    LO  & 98.9\scalebox{0.75}{$\pm$0.3} & 98.2\scalebox{0.75}{$\pm$0.4}
        & \underline{71.9\scalebox{0.75}{$\pm$0.5}}
        & 92.2\scalebox{0.75}{$\pm$0.7} & 90.4\scalebox{0.75}{$\pm$0.7}
        &\textbf{71.3\scalebox{0.75}{$\pm$0.9}}  \\
    KI  & 100.0\scalebox{0.75}{$\pm$0.0} & 99.9\scalebox{0.75}{$\pm$0.1} 
        & 95.9\scalebox{0.75}{$\pm$0.2}
        & \underline{81.9\scalebox{0.75}{$\pm$0.7}} & \textbf{79.9\scalebox{0.75}{$\pm$0.6}} 
        & 81.3\scalebox{0.75}{$\pm$0.7} \\

  \bottomrule
\end{tabular}}}
\label{tab:dis_results}
\end{table*}

\subsection{Discriminator Measure.} 
In order to verify whether the data we generated can be easily distinguished from the original data, we trained a LightGBM discriminator (with hyperparameter tuning) on a combination of the generated training set (with a label of 0) and the original training set (with a label of 1). Following this, we reported the test accuracy on a test data set, which comprises equal portions of samples from both the generated test set and the real test set. The scores, which are displayed in Table \ref{tab:dis_results}, demonstrate the superior performance of  ~\ours{}.

\subsection{Running Time}
\label{ap:time}
We analyze the running time of ~\ours{} and  baseline methods. 
The results of the
Adult Income dataset are in Table \ref{tab:time}. As can be seen from the results, the generation method based on language models requires more computational resources and time.

\begin{table}[!ht]
    \centering
    \caption{The running time in seconds on the Adult Income dataset of different methods in the privacy protection
    setting. The number of fine-tuning steps for GReaT, TapTap and ~\ours{} was 10k. A total of 36k samples were generated.}
    \begin{tabular}{l l l}
        \toprule
        & Training Time &Sampling Time\\
        \midrule
        \textbf{CTGAN} & 873 & 9 \\
        \hline
        \textbf{TVAE} & 360 & 3 \\
        \hline
        \textbf{TabDDPM} & 856 & 158 \\
        \hline
        \textbf{GReaT} & 960 & 895 \\
        \hline
        \textbf{Taptap} & 910 & 506 \\
        \hline
        \textbf{AIGT} & 906 &456\\
         \bottomrule
    \end{tabular}
    \label{tab:time}
\end{table}

\section{Hyperparameters Optimization}
\label{ap:tune_param}
We employ Optuna \cite{Akiba19kdd} for hyperparameter tuning of  LightGBM. For each specific dataset and model, we initially tune the model's hyperparameters using the original data. The determined set of hyperparameters is then consistently applied across all experiments on the dataset for all methods, ensuring a fair comparison.

\begin{table}[!ht]
    \centering
    \caption{Hyperparameter Space of  LightGBM.}
\resizebox{\linewidth}{!}{
\tiny
    \begin{tabular}{l l l}
        \toprule
        Models & Parameter & Values\\
        \midrule
        \multirow{6}{*}{LightGBM}   & learning\_rate & Uniform[0.01, 0.1] \\
                                    & num\_leaves  & Uniform[10, 100]\\
                                    & subsample & Uniform[0.5, 1.0]\\
                                    & colsample\_bytree & Uniform[0.5, 1.0] \\
                                    & min\_child\_samples & UniformInt[2, 100] \\
                                    &\#Iterations & UniformInt[100, 1000] \\
       
       \bottomrule
    \end{tabular}}
    \label{ap:hyper-params}
    \vspace{-0.5cm}
\end{table}

\end{document}